\documentclass{article}
\usepackage{arxiv}

\usepackage{geometry}
\usepackage{graphicx}
\usepackage[hidelinks]{hyperref}

\usepackage{amssymb}
\usepackage{bm}
\usepackage{lineno}
\usepackage{amsmath}

\usepackage{multirow}
\usepackage{float}
\usepackage{tabularx}
\usepackage{subcaption}
\usepackage{booktabs}
\usepackage{array,xcolor}

\usepackage{algorithm, algpseudocode}

\usepackage{cite}

\newtheorem{lemma}{Lemma}
\newenvironment{proof}{\paragraph{Proof:}}{\hfill$\square$}

\title{Distribution free uncertainty quantification in neuroscience-inspired deep operators}
\author{
  Shailesh Garg  \\
  Department of Applied Mechanics\\
  Indian Institute of Technology Delhi\\
  Hauz Khas, New Delhi 110016, India. \\
  \texttt{shailesh.garg@am.iitd.ac.in} \\
  \And
  Souvik Chakraborty  \\
  Department of Applied Mechanics\\
  Yardi School of Artificial Intelligence (YScAI)\\
  Indian Institute of Technology Delhi\\
  Hauz Khas, New Delhi 110016, India. \\
  \texttt{souvik@am.iitd.ac.in}}
\begin{document}
\maketitle
\begin{abstract}
Energy-efficient deep learning algorithms are essential for a sustainable future and feasible edge computing setups. Spiking neural networks (SNNs), inspired from neuroscience, are a positive step in the direction of achieving the required energy efficiency. However, in a bid to lower the energy requirements, accuracy is marginally sacrificed. Hence, predictions of such deep learning algorithms require an uncertainty measure that can inform users regarding the bounds of a certain output. 
In this paper, we introduce the Conformalized Randomized Prior Operator (CRP-O) framework that leverages Randomized Prior (RP) networks and Split Conformal Prediction (SCP) to quantify uncertainty in both conventional and spiking neural operators. 
To further enable zero-shot super-resolution in UQ, we propose an extension incorporating Gaussian Process Regression. 
This enhanced super-resolution-enabled CRP-O framework is integrated with the recently developed Variable Spiking Wavelet Neural Operator (VSWNO).
To test the performance of the obtained calibrated uncertainty bounds, we discuss four different examples covering both one-dimensional and two-dimensional partial differential equations.
Results demonstrate that the uncertainty bounds produced by the conformalized RP-VSWNO significantly enhance UQ estimates compared to vanilla RP-VSWNO, Quantile WNO (Q-WNO), and Conformalized Quantile WNO (CQ-WNO). These findings underscore the potential of the proposed approach for practical applications.
\end{abstract}

\section{Introduction}\label{section: introduction}
Machine learning \cite{jordan2015machine,mahesh2020machine,alpaydin2021machine} in the current day and age has crossed the initiation threshold and has steamrolled its way into various aspects of science and engineering \cite{dubourg2011reliability,roy2023support,thapa2024reliability,kobayashi2024explainable,kobayashi2024deep,sharma2022advances,garg2021machine,bai2024energy,wang2024artificial,bhaganagar2022novel,bhaganagar2024accelerated}. The increasing appearance of the term \textit{scientific machine learning} in literature \cite{cuomo2022scientific,thiyagalingam2022scientific,hey2020machine} is a testament to the same. A huge chunk of this credit goes to the rapid development of neural network \cite{gurney2018introduction,lecun2015deep,abiodun2018state} algorithms, which have shown tremendous performance across various scientific tasks. Developments like operator learning algorithms \cite{lu2021learning,li2020fourier,tripura2023wavelet,kovachki2023neural} have made it relatively easier to learn datasets originating from complex partial differential equations, which are often used to model the physical space around us, whether natural or artificial. However, despite the ongoing improvements, their performance is not at par with the gold standard techniques like finite element modeling \cite{felippa2004introduction,rao2010finite}. This is further true for spiking variants of these operator learning algorithms \cite{garg2024neuroscience,kahana2022spiking}, a domain that is particularly important from the view of a sustainable future and edge computing. 
%
Naturally, a confidence measure that can inform the user as to how reliable a specific output is and what its boundaries/extremes are becomes essential. Such measures can prove instrumental in tasks where some form of decision-making is involved. 

Uncertainty associated with the predictions/outputs of an underlying model is usually represented in the form of confidence intervals or uncertainty bounds. These bounds can give Conditional Coverage (CC) or Marginal CC (MCC) \cite{angelopoulos2023conformal,lei2018distribution} depending on the algorithm adopted for Uncertainty Quantification (UQ). To understand this, assume that we have a set of inputs with $n_s$ samples, and we have uncertainty bounds corresponding to each sample. Then CC implies that for all samples individually, the probability that the corresponding true value will lie within their stipulated bounds is $\geq 1-\alpha$, $\alpha \in (0,1)$ being the miscoverage level or error rate. MCC, on the other hand, is applicable to the whole population and states that among all $n_s$ samples's true values, $\geq n_s(1-\alpha)$ true values should lie between their uncertainty bounds. Although CC is more powerful than MCC, getting bounds that satisfy CC within neural network algorithms is non-trivial \cite{moya2024conformalized,ma2024calibrated}. Therefore, the focus is often on developing algorithms that yield MCC.
Unfortunately, this is often rendered ineffective because of the inherent assumptions associated with the algorithms. 
%
For example, in Bayesian approaches for UQ \cite{jospin2022hands,izmailov2021bayesian,blundell2015weight}, we often assume some prior distribution and make assumptions about the noise present in the system. In frequentist ensemble approaches as well, in order to report bounds that give coverage with probability $\geq 1-\alpha$, we have to take the Gaussian assumption. Such assumptions, coupled with imperfect optimization of the underlying model, lead to imperfect uncertainty bounds.

To tackle this challenge, researchers use UQ techniques like Conformal Prediction \cite{angelopoulos2023conformal,lei2018distribution,shafer2008tutorial} (CP). CP is a distribution-free UQ technique that provides bounds that satisfy MCC conditions without the underlying strong assumptions, as in the case of Bayesian approaches. In this paper, we propose a Conformalized Randomized Prior Operator (CRP-O) framework for UQ. The proposed framework utilizes the concept of Randomized Prior (RP) networks \cite{osband2018randomized} and extends it to a base neural operator in order to arrive at a heuristic measure of uncertainty. This initial estimate is then calibrated using Split CP \cite{angelopoulos2023conformal,lei2018distribution,shafer2008tutorial} to get the desired coverage using the estimated uncertainty and, hence, to satisfy the MCC condition. 
The motivation for adopting the RP networks to arrive at an initial heuristic estimate of uncertainty is that while being set in a deterministic framework, it also allows users to incorporate any prior information regarding the dataset, and it is trivial to implement even for complex architectures. 
We have further enhanced the CRP-O framework by exploiting a Gaussian Process to enable zero-shot super-resolution in UQ
The proposed enhanced CRP-O framework is used in conjunction with the Variable Spiking Wavelet Neural Operator \cite{garg2024neuroscience} (VSWNO), which is a spiking variant of vanilla Wavelet Neural Operator \cite{tripura2023wavelet} (WNO) architecture. The applicability of the CRP-O framework is also shown when vanilla WNO is used as the base operator. 

We note that there are two prior work that have proposed conformal prediction in neural operator. 
Moya \textit{et al.} \cite{moya2024conformalized} proposed a UQ technique for Deep Operator Network \cite{lu2021learning} (DeepONet) architecture using SCP, and it proposes the use of quantile regression for training the base operator. 
Ma \textit{et al.} \cite{ma2024calibrated} proposed Risk-Controlling Quantile Neural Operator (RCQNO)  thar relies on quntile regression for obtaining the initial uncertainty estimates and couples it with SCP. However, quantile regression often suffer from crossing quantiles, that can reduce the effectiveness of SCP. Furthermore, (RCQNO) \cite{ma2024calibrated} requires splitting the available datasets into three different parts, which can be problematic for applications with limited data.

Four examples are discussed to test the efficacy of the proposed approach, and we compare the coverage of uncertainty bounds obtained using the CRP-VSWNO/ CRP-WNO with those obtained using vanilla RP-VSWNO/ RP-WNO \cite{garg2023randomized}, Q-WNO, and conformalized Q-WNO (CQ-WNO). The results produced show that while the latter fails to satisfy the MCC condition, the former gives the required coverage for the test dataset when analyzed as a whole. A comparison of the performance of conformalized RP-WNO is also drawn against the performance of RCQNO. The highlights of the proposed framework include,
\begin{itemize}
    \item \textbf{Application to VSWNO}: In this paper, we show the applicability of the proposed framework to VSWNO. Spiking neural networks promise a more energy-efficient approach to deep learning, but because of their inherent sparse communication, the accuracy achieved is slightly affected. Thus, it becomes important to quantify the uncertainty associated with their predictions, and the proposed framework is able to capture the same effectively.
    \item \textbf{RP-Operator as the heuristic measure of uncertainty}: RP-Operator as an initial uncertainty estimator has the advantage of being set in a deterministic framework, which allows users to implement it for complex datasets easily. It can also incorporate prior information regarding the dataset, which is often omitted in a deterministic framework. It should be noted that despite its trivial nature, the underlying uncertainty estimates are robust enough that with the help of a meager calibration dataset, they can be improved to provide better coverage, and its mean predictions give a good approximation of the ground truth.
    \item \textbf{Limited data split}: The proposed framework requires the original dataset to be split into only two parts, (i) the training dataset and (ii) the calibration dataset. This is helpful in cases where the dataset available is sparse and limited.
    \item \textbf{Super-resolution predictions}: In this paper, using the Gaussian process \cite{williams2006gaussian,schulz2018tutorial}, we show how the proposed framework can be extended for uncertainty quantification in super-resolution predictions of RP operator networks.
\end{itemize}
The rest of the paper is arranged as follows. Section \ref{section: background} discusses the instances from literature where SCP is used for UQ in operator learning. Section \ref{section: proposed framework} details the proposed framework, and section \ref{section: numerical} shows the numerical illustrations. Section \ref{section: conclusion} concludes the findings of the paper.
\section{Related Works}\label{section: background}
In this section, we discuss the available literature on how the SCP algorithm is used for UQ in operator learning algorithms. The goal of the SCP algorithm is to provide uncertainty bounds that satisfy the MCC condition for a given trained network's output. Formally, given a calibration dataset having $n$ i.i.d. input-output pairs $(\bm u_i, y_i), \bm u_i\in\mathbb R^d, y_i \in\mathbb R$ and a test sample pair $(\bm u_t, y_{t})$, we want  a prediction band $\mathcal C_p(\bm u_t)$ that satisfies,
\begin{equation}
    \mathbb P(y_t\in \mathcal C_p(\bm u_t))\geq 1-\alpha,
\end{equation}
where the probability $\mathbb P(\cdot)$ is computed over the $n+1$ i.i.d pair of samples. To achieve this, we require a heuristic measure of uncertainty, which is then modified to obtain the required bands.
\subsection{Conformal Quantile DeepONet Regression}
Proposed in \cite{moya2024conformalized}, the conformal quantile DeepONet regression framework utilizes DeepONet \cite{lu2021learning} architecture to arrive at the base heuristic estimate of uncertainty and then utilizes SCP to calibrate the same and to provide the desired coverage. Two copies of the DeepONet model are trained on the same dataset using quantile regression \cite{hao2007quantile,koenker2001quantile}, one for $\alpha/2$ quantile and the other for $1-\alpha/2$ quantile. To train the network for $\lambda$ quantile, a pinball loss $L_p(y,y_{p_\lambda})$ function is used, and the same is defined as,
\begin{equation}
    L_p(y,y_{p_\lambda}) = \left\{\begin{matrix}
        \lambda(y-y_{p_\lambda}), & y-y_{p_\lambda}>0\\
        (1-\lambda)(y_{p_\lambda}-y), & y-y_{p_\lambda}\leq0
    \end{matrix}\right.,
\end{equation}
where $y$ is the true value and $y_{p_\lambda}$ is the model prediction when training for $\lambda$ quantile. Upon training the two base models, an initial confidence interval set is formed
The same is then calibrated using the SCP algorithm and a calibration dataset.
The results shown in \cite{moya2024conformalized} illustrate that, on an average, the framework yields the desired coverage; however, the coverage obtained falls short at a few locations in the solution's domain. Furthermore, quantile regression suffers from crossing quantiles, which is not addressed in \cite{moya2024conformalized}. As shown in the current paper, the quantile training for WNO architecture produces sub-par results that, while improved upon calibration, still fall short of providing desired coverage at all locations of the solution domain.
\subsection{Risk-Controlling Quantile Neural Operator}
The paper \cite{ma2024calibrated} proposes RCQNO to obtain ($\alpha$,$\delta$) risk-controlling prediction set that has two main components: (i) base operator and (ii) pre-calibration quantile neural operator. Fourier neural operator \cite{li2020fourier} is used as the background operator algorithm in this paper. The idea here is to first train the base operator using a training dataset $D_{tr_{I}}$. Afterward, using the trained base operator, residuals $|y-y_p|$ are computed corresponding to a different training dataset $D_{tr_{II}}$. The residuals and their corresponding inputs are used to train a new pre-calibration neural operator model with the goal of learning the point-wise radius of an assumed uncertainty ball corresponding to a given input function. Quantile regression is used to train this pre-calibration neural operator model. This point-wise radius forms the initial heuristic measure of uncertainty, which is then calibrated using a variant of SCP algorithm and a calibration dataset. 
This approach requires the dataset to be split into three portions, which is problematic for scenarios where the dataset is limited. Furthermore, the hyper-parameters used while calibrating the point-wise radius in the proposed algorithm of the paper greatly affect the amount of calibration dataset required, putting further strain on the already stretched dataset. 
\section{Proposed approach}\label{section: proposed framework}
In this section, we provide details of the proposed  Conformalized RP Operator (CRP-O) for enabling uncertainty quantification with guaranteed MCC. In particular, we have two technical contributions. First, we hypothesize that the initial uncertainty estimates have a significant effect on the performance of the SCP algorithm. To that end, we propose to use RP networks to estimate the base uncertainty. The advantages of RP networks over other alternatives such as Bayesian, quantile, and ensemble include their computational efficiency when parallelized, their scalability to complex neural network architectures, and their ability to account for prior information while being set in a deterministic framework. Second, one of the key features of neural operators is super-resolution. Here, we propose a GP based UQ framework that extends the CRP-O framework to enable UQ when predicting outputs at a resolution different from the one used during training. 
a UQ framework for spiking neural operators does not exist in the available literature.
We integrate the proposed framework with recent proposed VSWNO \cite{garg2024neuroscience} to illustrate how the proposed approach can enable UQ in neuroscience inspired operators. To our knowledge, this is the first attempt towards developing an UQ enabled neuroscience inspired neural operator.
We first briefly review VSWNO, a spiking variant of the vanilla WNO, and then proceed to introduce the CRP-O framework and, subsequently, the GP-based approach for enabling UQ in super-resolution.

%
\subsection{Variable spiking WNO (VSWNO)}
VSWNO \cite{garg2024neuroscience} is an energy-efficient variant of vanilla WNO (refer \ref{appendix:arch} for details on WNO) that exhibits event-driven computation behavior and sparse communication. To achieve this, it utilizes neuroscience-inspired Variable Spiking Neurons \cite{garg2023neuroscience} (VSNs) within its architecture. Specifically, the continuous activations $\sigma(\cdot)$ of artificial neurons used in vanilla WNO architecture are replaced by Variable Spiking Neurons \cite{garg2023neuroscience} (VSNs), giving rise to sparse communication and hence energy efficiency. The dynamics of a VSN is defined as follows, 
\begin{equation}
\begin{gathered}    
    \mathcal M_{t} = \beta_l \mathcal M_{t-1} + z_t,\\
    \widetilde y = \left\{ \begin{matrix}1, & \mathcal M_t\geq T_h\\
    0, & \mathcal M_t<T_h\end{matrix}\right.,\\
    \text{if } \widetilde y = 1, \mathcal M_t \leftarrow 0,\\
    y_t = \sigma(\widetilde y\,z_t), \text{ given } \sigma(0) = 0,
\end{gathered}
\end{equation}
where $y_t$ is the output of VSN corresponding to the input $z_t$ at the $t$\textsuperscript{th} time step of the input spike train. $\mathcal M_t$ represents the memory of the VSN at the $t$\textsuperscript{th} time step of the spike train. $\beta_l$ and $T_h$ are the leakage and threshold parameters of the VSN, which can treated as trainable parameters during network training. \cite{garg2024neuroscience} shows that VSWNO is able to produce results comparable to vanilla WNO with only a single time step in the spike train. Because the dynamics of VSN involve discontinuity in the form of hard threshold function, while training VSWNO, surrogate backpropagation \cite{neftci2019surrogate,garg2023neuroscience} is used. In surrogate backpropagation, during the backward pass of network training, the hard threshold function is idealized using an approximate smooth function, thus enabling gradient calculation. As discussed previously, we use the VSWNO in conjunction with RP networks to obtain our base estimate of the uncertainty in the proposed CRP-O framework. 

\subsection{Conformalized RP operator (CRP-O)}
The CRP-O framework comprises two key components: Randomized Prior (RP) networks and the Stochastic Cross-Validation Procedure (SCP) algorithm. The RP networks employ an ensemble-based training approach, enabling uncertainty quantification within a deterministic framework. This setup ensures straightforward implementation, even for complex deep learning architectures. Notably, RP networks extend beyond vanilla ensemble methods by incorporating prior information about the dataset under consideration.

The training process for RP networks involves optimizing \( n_c \) copies of the base network, each augmented with a prior network. Mathematically, let \( \mathcal{L}(\theta; \mathcal{D}) \) represent the loss function for the dataset \( \mathcal{D} \), where \( \theta \) denotes the network parameters. For each copy \( k \in \{1, 2, \ldots, n_c\} \), the augmented loss is defined as:

\begin{equation}
\mathcal{L}_{\text{aug}}^{(k)}(\theta^{(k)}; \mathcal{D}) = \mathcal{L}(\theta^{(k)}; \mathcal{D}) + \lambda \mathcal{L}_{\text{prior}}(\phi),
\end{equation}
where \( \mathcal{L}_{\text{prior}}(\phi) \) encodes prior knowledge through a non-trainable prior network with parameters \( \phi \), and \( \lambda \) is a weighting factor controlling the influence of the prior.
Each of the \( n_c \) augmented networks is initialized with different random seeds but trained on the same dataset \( \mathcal{D} \). The prior network, typically a neural network with an architecture similar to the base network, has fixed, non-trainable parameters. The incorporation of prior information can also take the form of known physics or domain-specific constraints.
This ensemble-based approach not only enhances robustness but also facilitates seamless integration of domain knowledge into the learning process, making RP networks a powerful tool for uncertainty-aware predictions.


In this paper, we use the term \textbf{RP operator} to signify the application of the RP network framework to neural operator algorithms. When the base operator network within the RP operator is chosen as the Variable-Width Spectral Neural Operator (VSWNO), the resulting model is denoted as \textbf{RP-VSWNO}. Similarly, if the base operator network is the standard Wavelet Neural Operator (WNO), the resulting model is referred to as \textbf{RP-WNO}. 
For RP-VSWNO, the prior network can either be an instance of VSWNO or a simpler WNO model. In this work, we use a smaller WNO model as the prior network for RP-VSWNO. Once the RP operator is trained, given an input \( \bm{u}_t \), the outputs \( y_{p(i)}(\bm{u}_t), \, i = 1, \dots, n_c \) are collected from all \( n_c \) copies of the operator network. The final mean \( \mu(\bm{u}_t) \) and standard deviation \( s(\bm{u}_t) \) are computed as:
\begin{subequations}\label{eq:rp_predict}
    \begin{equation}
        \mu(\bm{u}_t) = \dfrac{1}{n_c}\sum_{i=1}^{n_c} y_{p(i)}(\bm{u}_t),
    \end{equation}
    \begin{equation}
        s(\bm{u}_t) = \sqrt{\dfrac{1}{n_c} \sum_{i=1}^{n_c} \left(\mu(\bm{u}_t) - y_{p(i)}(\bm{u}_t)\right)^2}.
    \end{equation}
\end{subequations}
For non-scalar outputs, the above computations are performed element-wise for each dimension of the multi-dimensional output. Using the computed mean and standard deviation, the RP operator provides an initial confidence interval for a given input \( \bm{u}_t \) as:
\begin{equation}
    \mathcal{C}_{\text{ini}}(\bm{u}_t) = \left[ \mu(\bm{u}_t) - z s(\bm{u}_t), \, \mu(\bm{u}_t) + z s(\bm{u}_t) \right],
\end{equation}
where \( z \) is a parameter determined by the desired confidence level.
The initial uncertainty estimate \( \mathcal{C}_{\text{ini}} \) from the RP operator is further calibrated using the SCP algorithm. The final calibrated prediction band \( \mathcal{C}_p \) is defined as:
\begin{equation}
    \mathcal{C}_p(\bm{u}_t) = \left[ \mu(\bm{u}_t) - zq s(\bm{u}_t), \, \mu(\bm{u}_t) + zq s(\bm{u}_t) \right],
\end{equation}
where the parameter \( q \) is computed using the SCP algorithm. To determine \( q \), the trained RP operator is applied to a calibration dataset \( \{\bm{u}_i, y_i\}_{i=1}^n \) (distinct from the training dataset). For each calibration sample, the mean and standard deviation are obtained using Eq.~\eqref{eq:rp_predict}, and a score function \( e \) is computed as:
\begin{equation}
    e_i(\bm{u}_i, y_i) = \dfrac{\left| y_i - \mu(\bm{u}_i) \right|}{s(\bm{u}_i)}.
\end{equation}
For non-scalar outputs, the score function is calculated separately for each point in the output domain. The parameter \( q \) is then defined as the quantile of the score values over the calibration dataset:
\begin{equation}
    q = \text{Quantile}\left( \{ e_1, e_2, \dots, e_n \}, \frac{\lceil (1-\alpha)(n+1) \rceil}{n} \right),
\end{equation}
where \( \text{Quantile}(\mathcal{A}, a) \) represents the \( a \)-quantile of the set \( \mathcal{A} \), and \( \alpha \) corresponds to the desired confidence level. To ensure distinct score function values for the calibration dataset with high probability, jitter values can be added as necessary.

\begin{lemma}
The SCP algorithm provides calibrated prediction intervals such that for any arbitrary score function \( e(\bm{u}, y) \) that correctly estimates the discrepancy between predictions and true values, the probability of the true output \( y_t \) lying within the calibrated interval \( \mathcal{C}_p(\bm{u}_t) \) satisfies the guarantee:
\[
\mathbb{P}(y_t \in \mathcal{C}_p(\bm{u}_t)) \geq 1 - \alpha,
\]
for \( \alpha \geq \frac{1}{n+1} \), where \( n \) is the size of the calibration dataset.
\end{lemma}

\begin{proof}
Let the score function values \( \{e_i\}_{i=1}^n \) computed for the calibration dataset \( \{\bm{u}_i, y_i\}_{i=1}^n \) be sorted in increasing order as:
\begin{equation}
e_1 \leq e_2 \leq \dots \leq e_n.
\end{equation}
For a desired confidence level \( 1 - \alpha \), the parameter \( q \) is chosen as:
\begin{equation}
q = e_{\lceil (1-\alpha)(n+1) \rceil}.
\end{equation}
Given this choice of \( q \), the event \( y_t \in \mathcal{C}_p(\bm{u}_t) \) is equivalent to \( e_t \leq q \), i.e.,
\begin{equation}
    \{y_t \in \mathcal{C}_p(\bm{u}_t)\} = \{e_t \leq e_{\lceil (1-\alpha)(n+1) \rceil}\}.
\end{equation}
Under the assumption that the calibration samples \( \{(\bm{u}_i, y_i)\} \) are independent and identically distributed (i.i.d.), the distribution of \( e_t \) for a test sample \( (\bm{u}_t, y_t) \) matches that of the calibration scores \( \{e_i\}_{i=1}^n \). Thus, the probability of \( e_t \leq e_i \) is given by:
\begin{equation}
    \mathbb{P}(e_t \leq e_i) = \frac{i}{n+1}, \quad i \in \{1, 2, \dots, n\}.
\end{equation}
From this, the probability of \( e_t \leq e_{\lceil (1-\alpha)(n+1) \rceil} \) is:
\begin{equation}
\mathbb{P}(e_t \leq e_{\lceil (1-\alpha)(n+1) \rceil}) = \frac{\lceil (1-\alpha)(n+1) \rceil}{n+1}.
\end{equation}
By construction of \( \lceil \cdot \rceil \), it follows that:
\begin{equation}
\frac{\lceil (1-\alpha)(n+1) \rceil}{n+1} \geq 1 - \alpha.
\end{equation}
Therefore, the probability of \( y_t \) lying within the calibrated prediction interval is:
\begin{equation}
\mathbb{P}(y_t \in \mathcal{C}_p(\bm{u}_t)) = \mathbb{P}(e_t \leq e_{\lceil (1-\alpha)(n+1) \rceil}) \geq 1 - \alpha.
\end{equation}
For the special case where \( \alpha < \frac{1}{n+1} \), the choice \( q = \infty \) ensures that all \( e_t \leq q \), leading to \( \mathbb{P}(y_t \in \mathcal{C}_p(\bm{u}_t)) = 1 \). Hence, the guarantee is trivially satisfied.  For more details on SCP and its proof, readers may follow \cite{angelopoulos2023conformal}.
\end{proof}

\begin{figure}[ht!]
    \centering
    \includegraphics[width=0.85\textwidth]{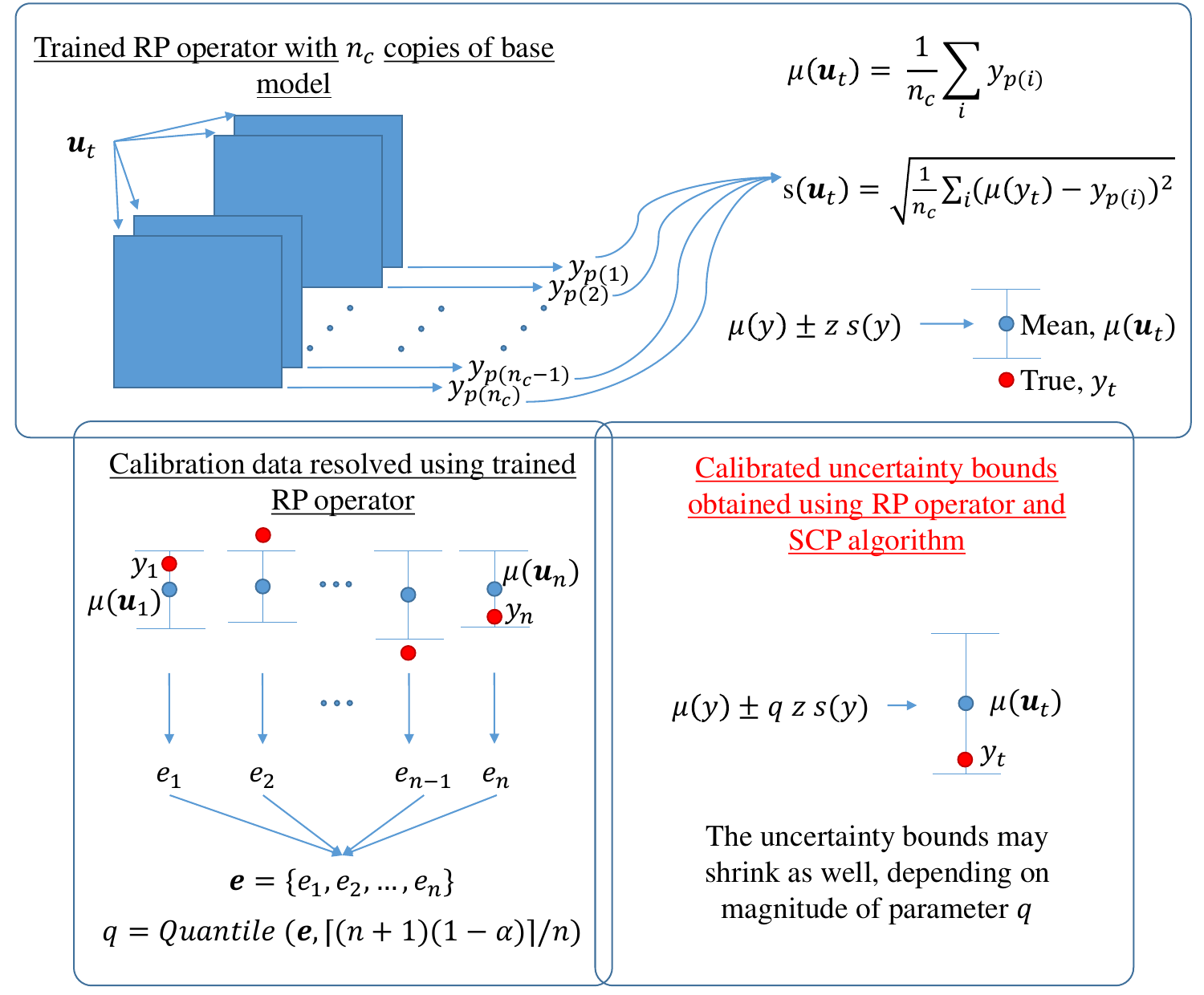}
    \caption{Schematic for the overall framework.}
    \label{fig:schematic}
\end{figure}

We note that VSWNO (or vanilla WNO), being an operator, processes multidimensional outputs obtained by discretizing the output function across spatial and temporal domains. Consequently, the SCP algorithm discussed earlier is applied element-wise to each element of the operator network's output.
For instance, consider a two-dimensional output where \( \mathbf{y} \in \mathbb{R}^{d_x \times d_y} \). In this case, the algorithm computes \( d_x \times d_y \) distinct values of the parameter \( q \), one for each element of the output matrix.
The detailed steps for computing the uncertainty bounds for multidimensional outputs are provided in Algorithm~\ref{algo}. Additionally, a schematic representation of the proposed framework is illustrated in Fig.~\ref{fig:schematic}.
\begin{algorithm}[ht!]
\caption{Algorithm for computing uncertainty set $\mathcal C_p$.}
\label{algo}
\begin{algorithmic}[1]
\Require Calibration dataset $D_C=\{\bm u_i, \bm y_i\}_{i=1}^{n}$, trained RP-WNO/ RP-VSWNO model, test input $\bm u_t$.
\State Obtain $\bm\mu(\bm u_i)$ and $\bm s(\bm u_i)$ using trained RP-WNO/ RP-VSWNO for each sample in the calibration dataset.
\State Compute score function $\bm e_i$ for each sample of calibration dataset, using $\bm e_i=|\bm y_i-\bm \mu(\bm u_i)|\oslash \bm s_i$, where $\oslash$ represents element-wise division.
\For{$j=1$ : no. of discretizations of solution grid}
\State Prepare a vector $\bm v$ containing magnitudes of $j$\textsuperscript{th} element of score functions $\bm e_i, i=1,\dots,n$. 
\State Conformal parameter $q_j$ corresponding to $j$\textsuperscript{th} element of output is equal to $\lceil(1-\alpha)(n+1)\rceil/n$\textsuperscript{th} quantile of vector $\bm v$.
\EndFor
\State $\bm q = \{q_1,q_2,\dots,q_{n_e}\}$, where $n_e=$ numbers of elements in the solution grid.
\State Reshape $\bm q$ to have the same shape as outputs $\bm \mu(\bm u_t)$ (or $\bm s(\bm u_t)$).
\State $\mathcal C_p = \left[\bm\mu(\bm u_t)-\bm s(\bm u_t)z\circ \bm q,\,\,\bm \mu(\bm u_t)+\bm s(\bm u_t)z\circ \bm q\right]$, where $\circ$ represents elementwise multiplication.
\renewcommand{\algorithmicrequire}{\textbf{Output:}}
\Require{Calibrated uncertainty set $\mathcal C_p(\bm u_t)$.}
\end{algorithmic}
\end{algorithm}
\subsection{Super-resolution predictions}
A key feature of operator learning algorithms is their ability to predict outputs at resolutions different from the training resolution. However, such super-resolution predictions are inherently more error-prone, making it essential to quantify the associated uncertainty. To extend the proposed UQ framework in super-resolution setup, we leverage a Gaussian Process (GP) model \cite{schulz2018tutorial, chakraborty2019graph}.
First, the parameter \( q_i \) is computed for each location \( \bm{x}_{\text{tr}_i} \in \mathbb{R}^d \) on the initial solution grid \( \mathcal{G}_{\text{tr}} \) (refer to Algorithm~\ref{algo}). These computed values are then mapped to the corresponding locations \( \bm{x}_{\text{sr}_i} \in \mathbb{R}^r, r>d \) on the super-resolution solution grid \( \mathcal{G}_{\text{sr}} \) using the Gaussian Process.

Given a function \( q = f(\bm{x}) \) with $q$ being the conformal parameter, a Gaussian Process (GP) defines a distribution over this function as:
\begin{equation}
    f(\bm{x}) \sim GP(M(\bm{x}; \bm{\theta}_M), K(\bm{x}, \bm{x}'; \bm{\theta}_K)),
\end{equation}
where \( M(\bm{x}; \bm{\theta}_M) \) and \( K(\bm{x}, \bm{x}'; \bm{\theta}_K) \) denote the mean function and covariance function, respectively. The parameters of these functions are represented by \( \bm{\theta}_M \) and \( \bm{\theta}_K \).
Considering zero-mean Gaussian Process, we have:
\begin{equation}\label{eq:gp_zero}
    f(\bm{x}) \sim GP(0, K(\bm{x}, \bm{x}'; \bm{\theta}_K)).
\end{equation}
Given a set of input locations \( \bm{\hat{x}} = \{\bm{x}_1, \dots, \bm{x}_d\} \) and corresponding outputs \( \bm{q} = \{q_1, \dots, q_d\} \), the GP defined in Eq. \eqref{eq:gp_zero} reduces to 
\begin{equation}
    p(f | \bm{\hat{x}}) \sim \mathcal{N}(f | \bm{0}, \mathbf{K}),
\end{equation}
where \( \mathbf{K} \) represents the covariance matrix. Note that the hyperparameters, $\bm \theta_K$ of $\mathbf K$ includes the length-scale parameter and the process variance, which are obtained by maximizing the marginal likelihood (or minimizing the negative log marginal likelihood). Details on training GP is shown in Algorithm~\ref{algo: GP}). 
Once trained, the predictions at a  new location \( \bm{x}^* \) can be obtained 
by computing the predictive distribution, \( p(q^* | \bm{x}^*, \bm{\hat{x}}, \bm{q}) \),
\begin{equation}
    p(q^* | \bm{x}^*, \bm{\hat{x}}, \bm{q}) \sim \mathcal{N}(q^* | \mu^*, K^*),
\end{equation}
where 
\begin{equation}
    \mu^* = \mathbf{K}(\bm{\hat{x}}, \bm{x}^*)^\top \mathbf{K}(\bm{\hat{x}}, \bm{\hat{x}}')^{-1}\bm q
\end{equation}
and
\begin{equation}
    K^* = K(\bm{x}^*, \bm{x}^*) - \mathbf{K}(\bm{\hat{x}}, \bm{x}^*)^\top \mathbf{K}(\bm{\hat{x}}, \bm{\hat{x}}')^{-1} \mathbf{K}(\bm{\hat{x}}, \bm{x}^*).
\end{equation}
Here, \( K(\bm{x}^*, \bm{x}^*) \in \mathbb{R} \), \( \mathbf{K}(\bm{\hat{x}}, \bm{x}^*) \in \mathbb{R}^{n \times 1} \), and \( \mathbf{K}(\bm{\hat{x}}, \bm{\hat{x}}') \in \mathbb{R}^{n \times n} \). 
Using this procedure, the GP is trained to model the parameter \( q \) over the locations \( \bm{x}_{\text{tr}_i} \) on the training grid \( \mathcal{G}_{\text{tr}} \) and subsequently predict \( q \) at the super-resolution grid locations \( \bm{x}_{\text{sr}_i} \) on \( \mathcal{G}_{\text{sr}} \). We utilize the predictive mean obtained from GP to obtain the calibrated uncertainty.

One key aspect of GP is selection of covariance kernel. In this work, a rational quadratic function is used as the covariance kernel, and the GP is modeled and trained using the \texttt{GaussianProcessRegressor} function from Python’s \texttt{Scikit-Learn}\footnote{\url{https://scikit-learn.org/stable/modules/generated/sklearn.gaussian_process.GaussianProcessRegressor.html}} library.

\begin{algorithm}[ht!]
\caption{Algorithm for training GP.}
\label{algo: GP}
\begin{algorithmic}[1]
\Require Training dataset $\{x_{tr_i},q_i\}_{i=1}^{n_e}$, covariance function $K(\cdot,\cdot;\bm\theta_K)$, initial estimates for $\bm\theta_K$, $\bm\theta_{K_0}$, error threshold $\epsilon$, and an optimizer algorithm.
\State $\bm\theta_K\leftarrow\bm\theta_{K_0}$, $\epsilon'\leftarrow10\,\epsilon$
\While {$\epsilon'>\epsilon$}
\State Compute the negative log-likelihood $L_{nll}$ using training data and parameters $\bm\theta_K$.
\State $L_{nll} \propto -\dfrac{1}{2}(\log |\mathbf K|+\bm y^T\mathbf K^{-1}\bm y)$, where $\bm y$ represents the observation vector.
\State Update parameters $\theta_K$ using suitable optimizer.
\State $\bm\theta_{K_{new}}\leftarrow\bm\theta_K-g\left(\dfrac{\partial L_{nll}}{\partial\bm\theta_K}\right)$, where $g(\cdot)$ is governed by the choice of optimizer.
\State $\bm\theta_{K_{old}}\leftarrow\bm\theta_K$
\State $\bm\theta_K\leftarrow\bm\theta_{K_{new}}$
\State $\epsilon' = ||\bm\theta_{K_{new}}-\bm\theta_{K_{old}}||_2^2$
\EndWhile
\renewcommand{\algorithmicrequire}{\textbf{Output:}}
\Require{Optimized GP parameters $\bm\theta_K$.}
\end{algorithmic}
\end{algorithm}
\section{Numerical Illustrations}\label{section: numerical}
In this section, we present four examples to illustrate the efficacy of the proposed framework. 
While the first example
deals with the Burger's equation, the second and third examples deal with 
the two-dimensional Darcy equation defined on a rectangular domain and triangular domain, respectively.
As the fourth example, the seismic wave propagation is considered. 
For all the examples, we present results obtained using both CRP-WNO and CRP-VSWNO models, and compare the results obtained with those obtained using 
RP-WNO, RP-VSWNO, Q-WNO, and CRQ-WNO. 
Additionally, for the fourth example, we also present a comparative assessment between CRP-WNO and RCQNO models. 

The first example deals with a dataset generated using the one-dimensional Burgers equation, the second and third examples deal with the two-dimensional Darcy equation defined on a rectangular domain and triangular domain, respectively, and the fourth example deals with a dataset generated using the Helmholtz equation. To learn the datasets in different examples, we use the CRP-WNO and CRP-VSWNO models and compare the results with those obtained when using RP-WNO, RP-VSWNO, Q-WNO, and CRQ-WNO. In the fourth example, we also compare the performance of CRP-WNO against that of RCQNO model.
The nomenclature for various deep learning models mentioned above are shown in Table \ref{tab:Nomenclature}.
\begin{table}[ht!]
    \centering
    \caption{Nomenclature for various deep learning models used in the four examples.}
    \vspace{0.5em}
    \label{tab:Nomenclature}    
    \begin{tabular}{ll}
         \toprule
         Model& Description \\
         \midrule
         RP-WNO & Randomized Prior Wavelet Neural Operator\\
         CRP-WNO & Conformalized (calibrated) RP-WNO \\
         RP-VSWNO & Randomized Prior Variable Spiking Wavelet Neural Operator \\
         CRP-VSWNO & Conformalized RP-VSWNO \\
         Q-WNO & Quantile Wavelet Neural Operator \\
         CQ-WNO & Conformalized Q-WNO \\
         \bottomrule
    \end{tabular}
\end{table}
Individual architecture details for WNO based deep learning models in various examples are given in \ref{appendix:arch}. For details on the training of Q-WNO and calibration process of CQ-WNO, refer  \ref{appendix:qwno}. The quantum of training data, calibration data, and the testing data used in various examples is given in Table \ref{tab:dataquantum}. It should be noted that while we are demonstrating results using only 100 test samples in various examples, trained deep learning models have no such limitation and can be used any number of test samples.
\begin{table}[ht!]
    \centering
    \caption{Dataset size used in various examples.}
    \vspace{0.5em}
    \label{tab:dataquantum}
    \begin{tabular}{lccc}
    \toprule
    \multirow{2}{*}{Example} & \multicolumn{3}{c}{No. of Samples in Dataset} \\
    \cmidrule{2-4}
     & Training & Calibration & Testing\\
     \midrule
     E-I, Burgers Equation & 1000 & 50 & 100\\
     E-II, Darcy Equation - Rectangular Domain & 800 & 100 & 100\\
     E-III, Darcy Equation - Triangular Domain & 1700 & 150 & 100\\
     E-IV, Helmholtz Equation & 500 & 150 & 100\\
     \bottomrule
    \end{tabular}
\end{table}
For all the examples, we aim to have $95\%$ coverage and hence set $\alpha = 0.05$. The coverage reported in various examples, corresponding to various deep learning models, is based on the observations of the test dataset. Samples from the calibration dataset are not included in coverage results to maintain fairness in comparison. For all the examples, we consider an ensemble size of 10 for the RP operator networks.
Finally, one of the main motivations behind neuroscience inspired operators is to reduce energy consumption.
However, the potential of the spiking neurons and its variants such as the variable spiking neurons in energy saving can only be realized while using neuromorphic hardware.
In absence of availability of such  hardware, a common
surrogate metric for energy measurement is the spiking activity \cite{garg2024neuroscience,kahana2022spiking}.
In this study also, we use the same metric to evaluate energy consumption, and report the results obtained in  
\ref{appendix:spkact}.


\subsection{E-I, Burgers Equation}
In the first example, we consider the one-dimensional Burgers equation, which is a nonlinear PDE and is often used in fluid dynamics to model the flow of a viscous fluid. It also finds its application in fields like gas dynamics, nonlinear acoustics, and traffic flow. The Burgers equation is defined as,
\begin{equation}\label{eq:burger}
    u_t(x,t)+u(x,t)u_x(x,t) = \nu\,u_{xx}(x,t),\,\,x\in [0,1], t\in (0,1],
\end{equation}
where $\nu=0.1$ is the viscosity coefficient. Periodic boundary conditions of the form $u(x=0,t)=u(x=1,t)$ are considered. Our objective is to learn the operator that maps from initial condition $u(x,0)$ to the solution at $t=1$, i.e., $u(x,1)$,
\[ \mathcal M: u(x,0) \mapsto u(x,1).\]
%
We assume $u(x,0)$ to be a Gaussian random field, 
\begin{equation}
    u(x,0)\sim N\left(\bm 0,625(-\Delta+25\mathbb I)^{-2}\right),
\end{equation}
where $\Delta$ is the Laplacian.
For generating data, Eq. \eqref{eq:burger} is solved using $\Delta t = 1/200$ and by discretizing the spatial domain into 8192 grid points. Once generated, the data is subsampled into a grid with $1024$ grid points. For more information on the dataset, readers may follow the source for the dataset, \cite{li2020fourier}.


\begin{figure}[ht!]
    \centering
    \begin{subfigure}{0.49\textwidth}
    \centering
    \includegraphics[width=\textwidth]{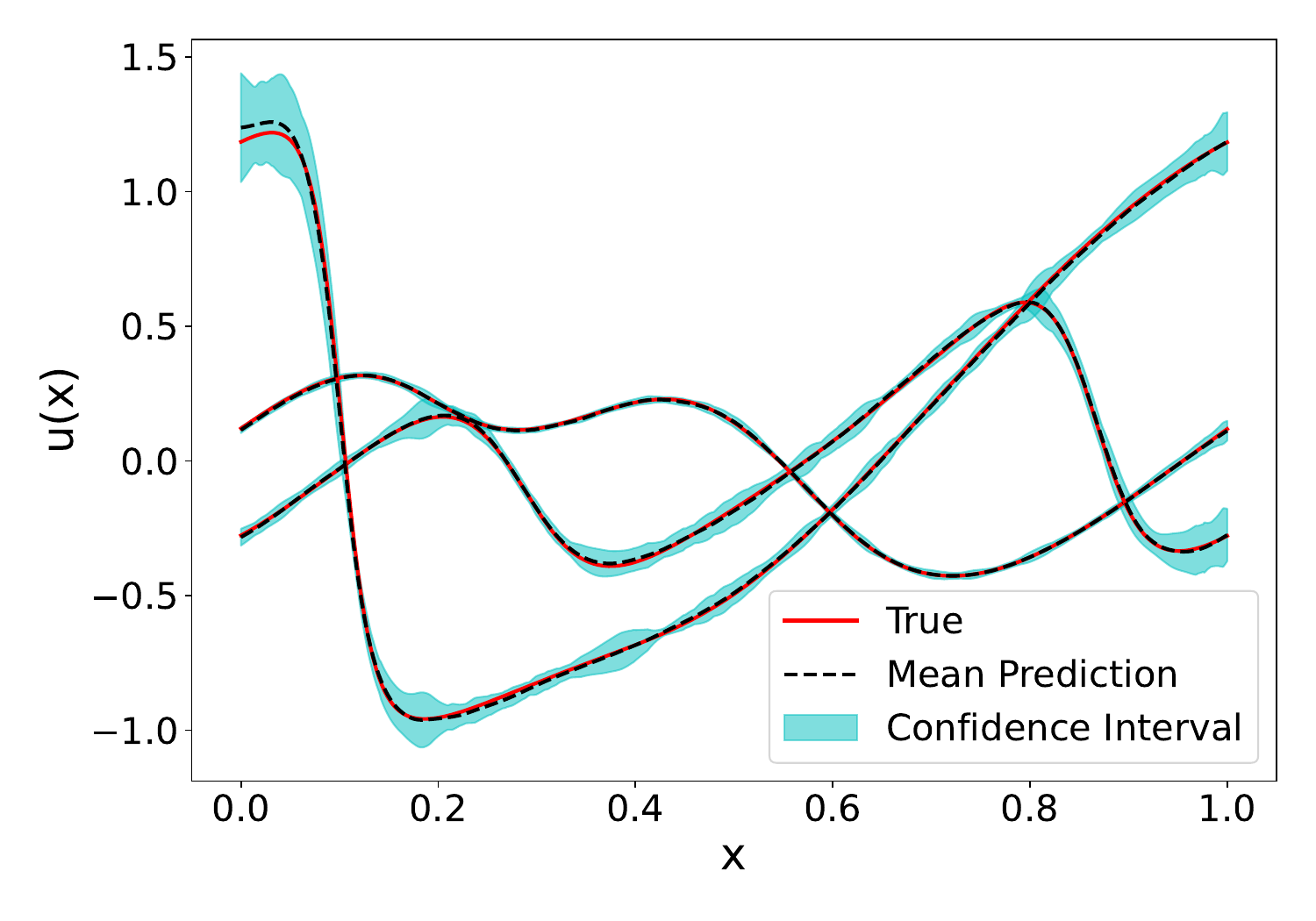}
    \caption{CRP-WNO, Normalized Mean Square Error (NMSE) = 0.046\%.}
    \end{subfigure}
    \begin{subfigure}{0.49\textwidth}
    \centering
    \includegraphics[width=\textwidth]{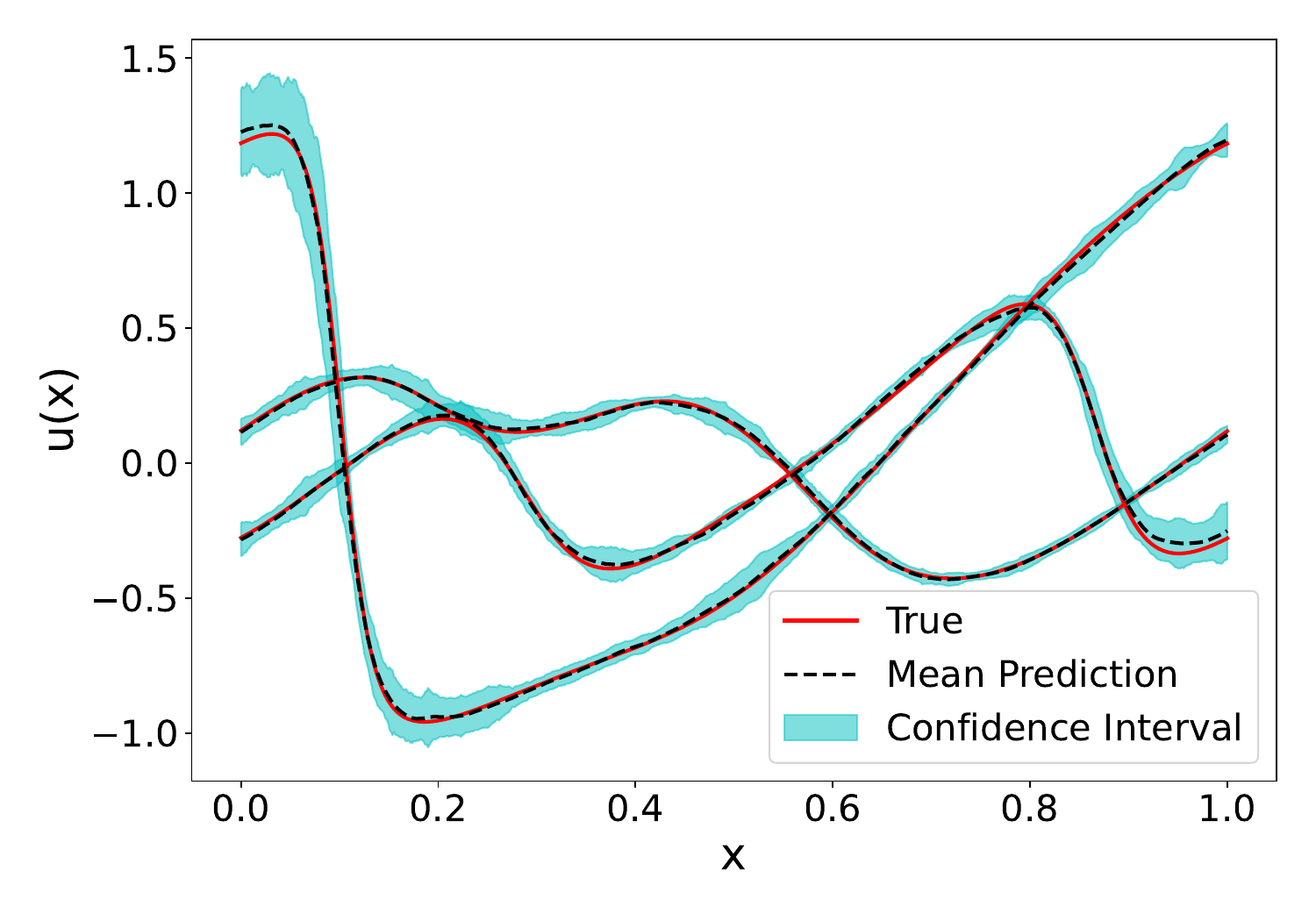}
    \caption{CRP-VSWNO, NMSE = 0.085\%.}
    \end{subfigure}
    \caption{Mean predictions for test inputs obtained using CRP-WNO and CRP-VSWNO compared against the ground truth in E-I. Calibrated confidence intervals obtained using the two networks are also plotted for different test samples.}  \label{fig:burgers_conformal_predictions}
\end{figure}
Fig. \ref{fig:burgers_conformal_predictions} shows the mean predictions of CRP-WNO and CRP-VSWNO networks for different test samples and their respective calibrated confidence intervals. The mean predictions closely follow the ground truth, and they are engulfed by the confidence intervals in the regions where they diverge.
\begin{table}[ht!]
    \centering
    \caption{Coverage provided by the confidence intervals generated using various deep learning models and their conformalized versions in E-I. Count for the number of spatial locations where coverage is $<95\%$ or is $\geq 95\%$ is also included.} 
    \vspace{0.5em}
    \label{tab:cover_burgers}
    \begin{tabular}{lccccc}
    \toprule
    \multirow{2}{*}{Model} & \multicolumn{3}{c}{Coverage} & \multicolumn{2}{c}{Count (Out of 1024)} \\
    \cmidrule(l{2pt}r{2pt}){2-4}\cmidrule(l{2pt}r{2pt}){5-6}
    & Average & Min. & Max. & $<$ 95\% & $\geq$ 95\%\\
    \midrule
    RP-WNO&94.23 & 82.00 & 100.00 & 422 & 602\\
    \textbf{CRP-WNO} & \textbf{99.89} & \textbf{98.00} & \textbf{100.00} & \textbf{0} & \textbf{1024}\\
    RP-VSWNO & 97.42 & 92.00 & 100.00 & 82 & 942\\
    \textbf{CRP-VSWNO} & \textbf{99.66} & \textbf{96.00} & \textbf{100.00} & \textbf{0} & \textbf{1024}\\
    Q-WNO & 15.83 & 5.00 & 25.00 & 1024 & 0\\
    CQ-WNO & 96.32 & 89.00 & 100.00 & 265 & 759\\
    \bottomrule
     & 
    \end{tabular}
\end{table}
Table \ref{tab:cover_burgers} shows the coverage obtained corresponding to various deep learning models. As can be seen that the RP operators give a decent initial measure and the proposed framework utilizing the same RP operator predictions is able to give the desired 95\% coverage at all spatial locations. Furthermore, as can be observed, the performance of CQ-WNO is lacking severely compared to the performance of CRP-WNO or CRP-VSWNO. While CQ-WNO is able to give a coverage of $>95\%$ when average is taken over all spatial locations, when looked at individually, at several locations, its coverage is $<95\%$. This is not the case with either of the CRP networks. 

\begin{figure}[ht!]
    \centering
    \begin{subfigure}{0.49\textwidth}
    \centering
    \includegraphics[width=0.8\textwidth]{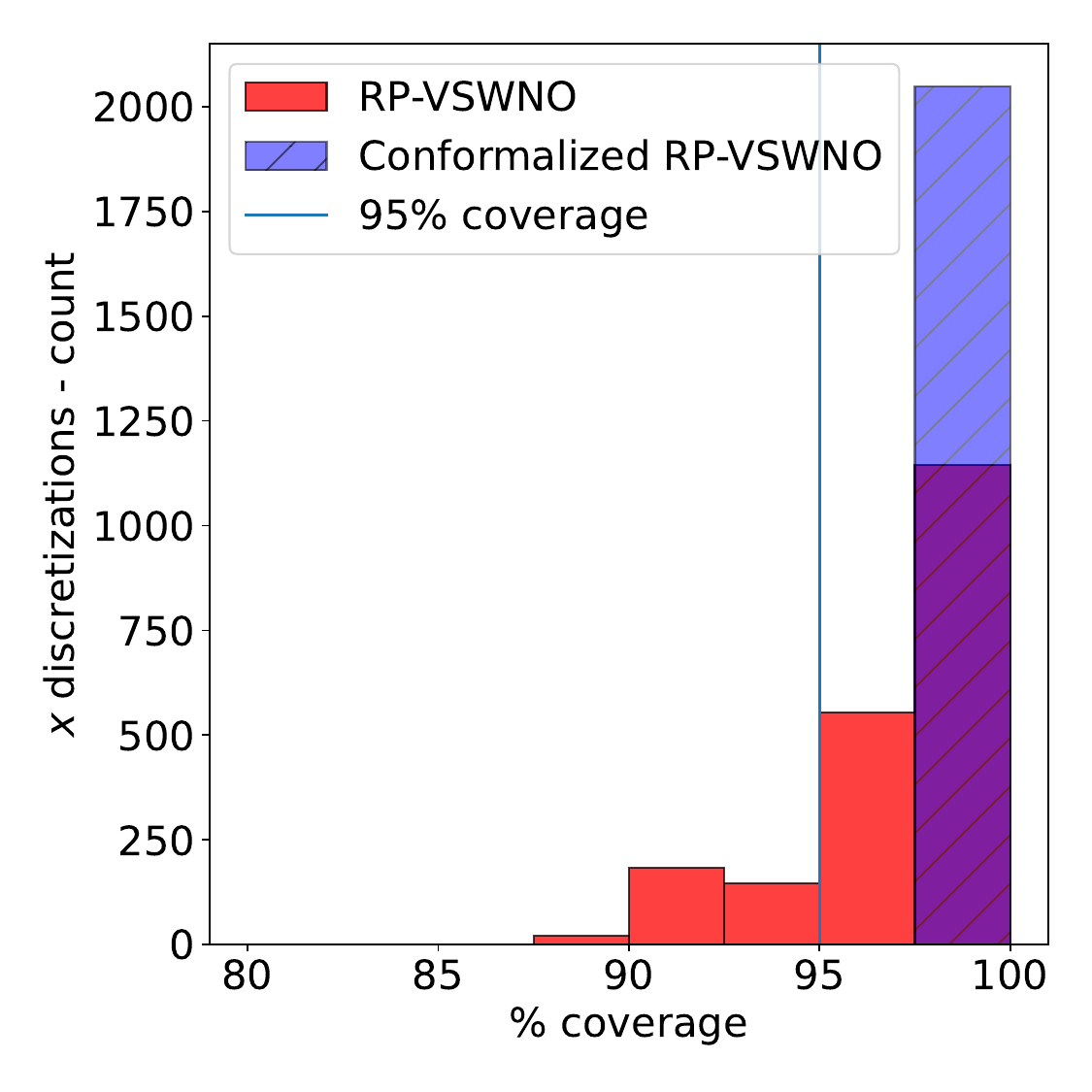}
    \caption{Coverage provided by the confidence intervals.}
    \end{subfigure}
    \begin{subfigure}{0.49\textwidth}
    \centering
    \includegraphics[width=\textwidth]{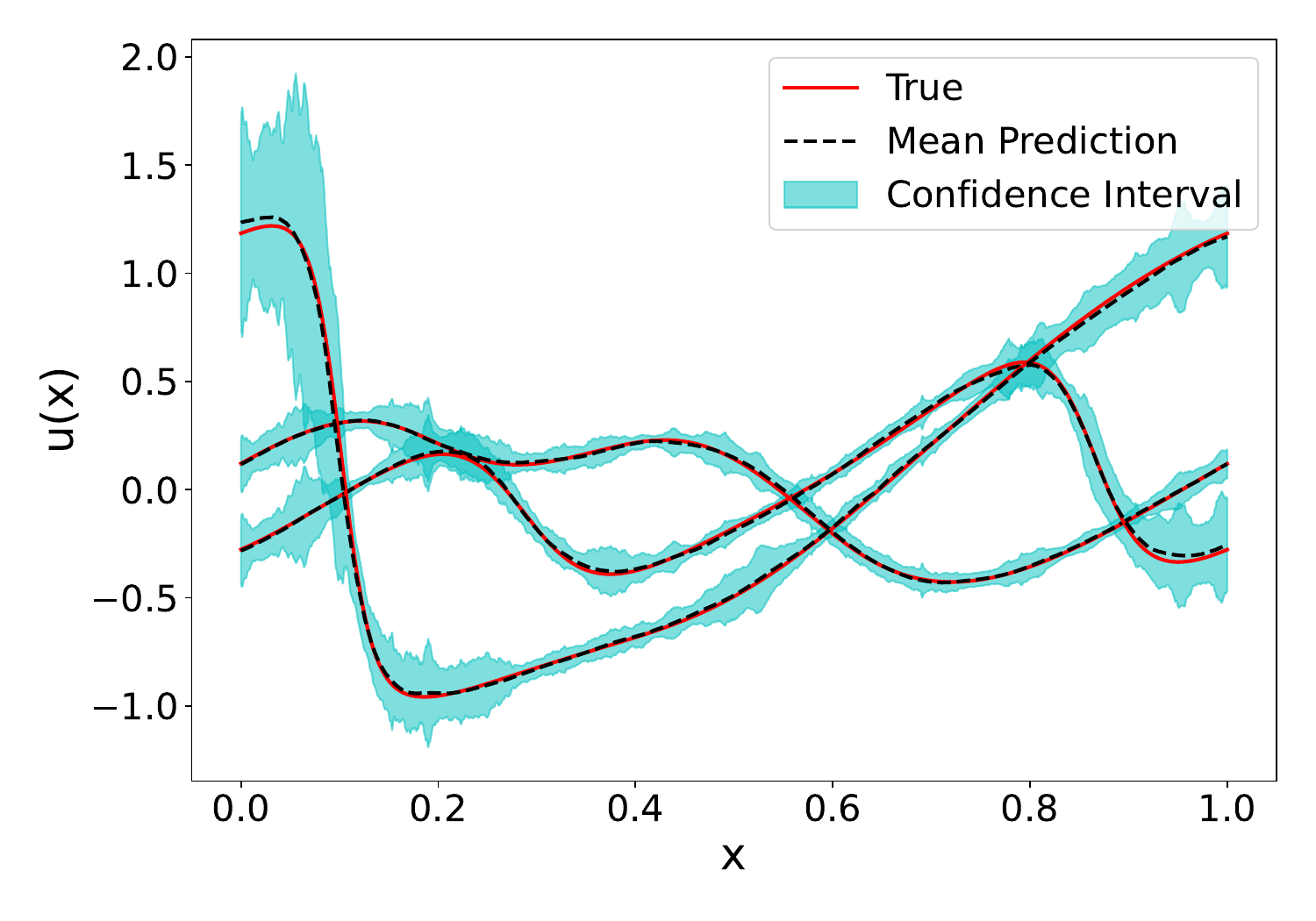}
    \caption{Mean predictions compared against ground truth and confidence intervals obtained using CRP-VSWNO. NMSE = 0.086\%.}
    \end{subfigure}
    \caption{Super-resolution results for E-I.}
    \label{fig:burgers_super}
\end{figure}
Fig. \ref{fig:burgers_super}(a) shows the coverage obtained for RP-VSWNO and CRP-VSWNO when predicting outputs at a higher resolution. In the current example, we have trained and calibrated our models for a spatial resolution of 1024. However, for our super-resolution results, we predict outputs corresponding to a spatial resolution of 2048. As can be seen from Fig. \ref{fig:burgers_super}(a), the coverage improves significantly (as opposed to initial uncertainty estimates obtained from the RP operator) despite the original calibration being done on a coarser grid. Fig. \ref{fig:burgers_super}(b) shows the mean predictions of CRP-VSWNO compared against the ground truth and the corresponding confidence intervals when predicting at the higher resolution of 2048. The mean predictions closely follow the ground truth, and the confidence intervals perform their function well. It should be noted here that the width of confidence intervals obtained when predicting at a higher resolution is more than the width of confidence intervals obtained when predicting at the training resolution. This is because the model is now also predicting at locations that were not considered during training.
\subsection{E-II, Darcy Equation on Rectangular Domain}
The second example deals with a two-dimensional Darcy flow equation often used in literature to model flow through a porous media, 
\begin{equation}
    -\nabla(a(x,y)\nabla u(x,y)) = 1,\,\,x,y\in(0,1),
\end{equation}
where $a(x,y)$ is the permeability and $u(x,y)$ is the pressure. Zero Dirichlet boundary conditions are considered while generating the dataset. The operator to be learned in this example maps the permeability field $a(x,y)$ to the pressure field $u(x,y)$,
\begin{equation}
    \mathcal M: a(x,y) \mapsto u(x,y).
\end{equation}
A spatial resolution of $85\times 85$ is considered, resulting in $\delta x = \delta y = 1/85$. The permeability field is assumed to be a Gaussian random field defined as
\begin{equation}
  a(x,y) \sim N\left(\bm 0,(-\Delta+9\mathbb I)^{-2}\right),
\end{equation}
which is then passed through a mapping function which sets all negative values equal to 3 and all positive values equal to 12. For more details regarding the dataset, refer \cite{li2020fourier,tripura2023wavelet}. 

\begin{figure}[ht!]
    \centering
    \begin{subfigure}{0.8\textwidth}
    \centering
    \includegraphics[width=\textwidth]{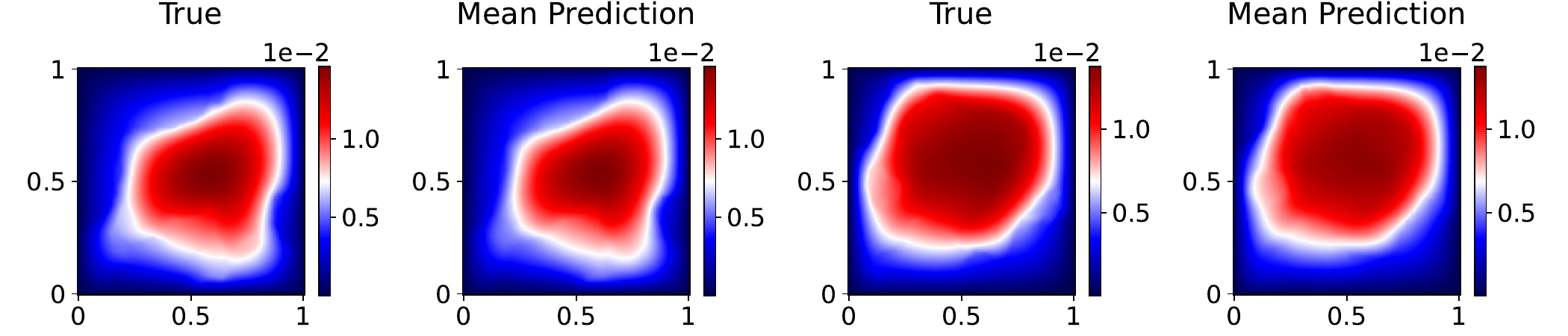}
    \caption{CRP-WNO, NMSE = 0.039\%.}
    \end{subfigure}
    \begin{subfigure}{0.8\textwidth}
    \centering
    \includegraphics[width=\textwidth]{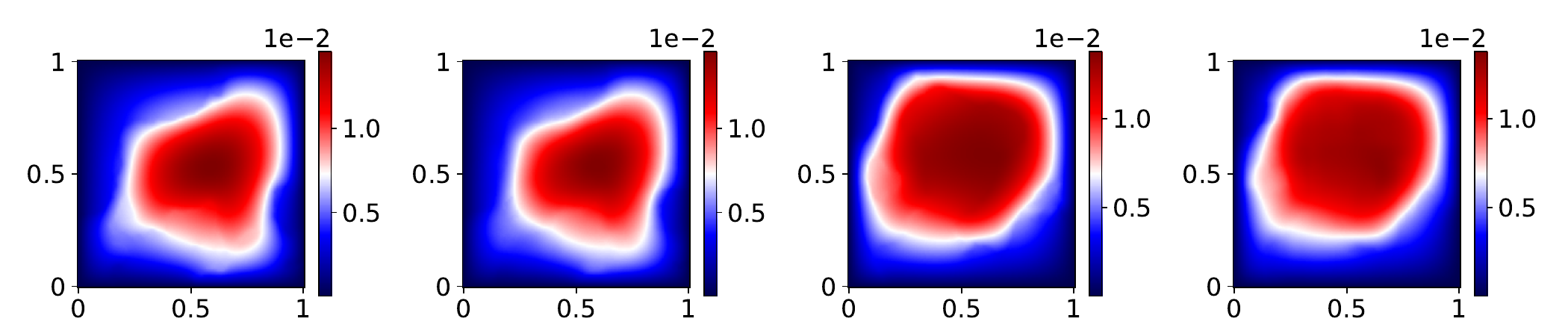}
    \caption{CRP-VSWNO, NMSE = 0.044\%.}
    \end{subfigure}
    \caption{Mean predictions for test inputs obtained using CRP-WNO and CRP-VSWNO compared against the ground truth in E-II.}
    \label{fig:Darcytruevspred}
\end{figure}
Fig. \ref{fig:Darcytruevspred} shows the mean predictions from CRP-WNO and CRP-VSWNO compared against the ground truth. The mean predictions give a good approximation of the ground truth. Table \ref{tab:CoverageDarcyRect} shows the performance of confidence intervals obtained using various deep learning models. 
\begin{table}[ht!]
    \centering
    \caption{Coverage provided by the confidence intervals generated using various deep learning models and their conformalized versions in E-II. Count for the number of discretizations of $(x,y)$ grid where coverage is $< 95\%$ or is $\geq 95\%$ is also included.}
    \vspace{0.5em}
    \label{tab:CoverageDarcyRect}
    \begin{tabular}{lccccc}
    \toprule
    \multirow{2}{*}{Model} & \multicolumn{3}{c}{Coverage} & \multicolumn{2}{c}{Count (Out of 7225)} \\
    \cmidrule(l{2pt}r{2pt}){2-4}\cmidrule(l{2pt}r{2pt}){5-6}
    & Average & Min. & Max. & $<$ 95\% & $\geq$ 95\%\\
    \midrule
    RP-WNO& 72.15 & 4.00 & 95.00 & 7224 & 1\\
    \textbf{CRP-WNO} & \textbf{99.49} & \textbf{95.00} & \textbf{100.00} & \textbf{0} & \textbf{7225}\\
    RP-VSWNO & 83.12 & 4.00 & 100.00 & 6413 & 812\\
    \textbf{CRP-VSWNO} & \textbf{99.62} & \textbf{95.00} & \textbf{100.00} & \textbf{0} & \textbf{7225}\\
    Q-WNO & 51.26 & 0.00 & 94.00 & 7225 & 0\\
    CQ-WNO & 95.07 & 81.00 & 100.00 & 2668 & 4557\\
    \bottomrule
     & 
    \end{tabular}
\end{table}
In this example, the performance of RP-WNO and RP-VSWNO is subpar. However, their conformalized versions, CRP-WNO and CRP-VSWNO, give the expected coverage of $\geq 95\%$ at all locations of the $(x,y)$ grid. Furthermore, the performance of CQ-WNO is worse than that of both the conformalized RP operators. Similar to the previous example, although CQ-WNO reaches the $95\%$ mark in the average sense, it fails to provide adequate coverage at several locations of $(x,y)$ grid.

\begin{figure}[ht!]
    \centering
    \begin{subfigure}{0.55\textwidth}
    \centering
    \includegraphics[width=0.8\textwidth]{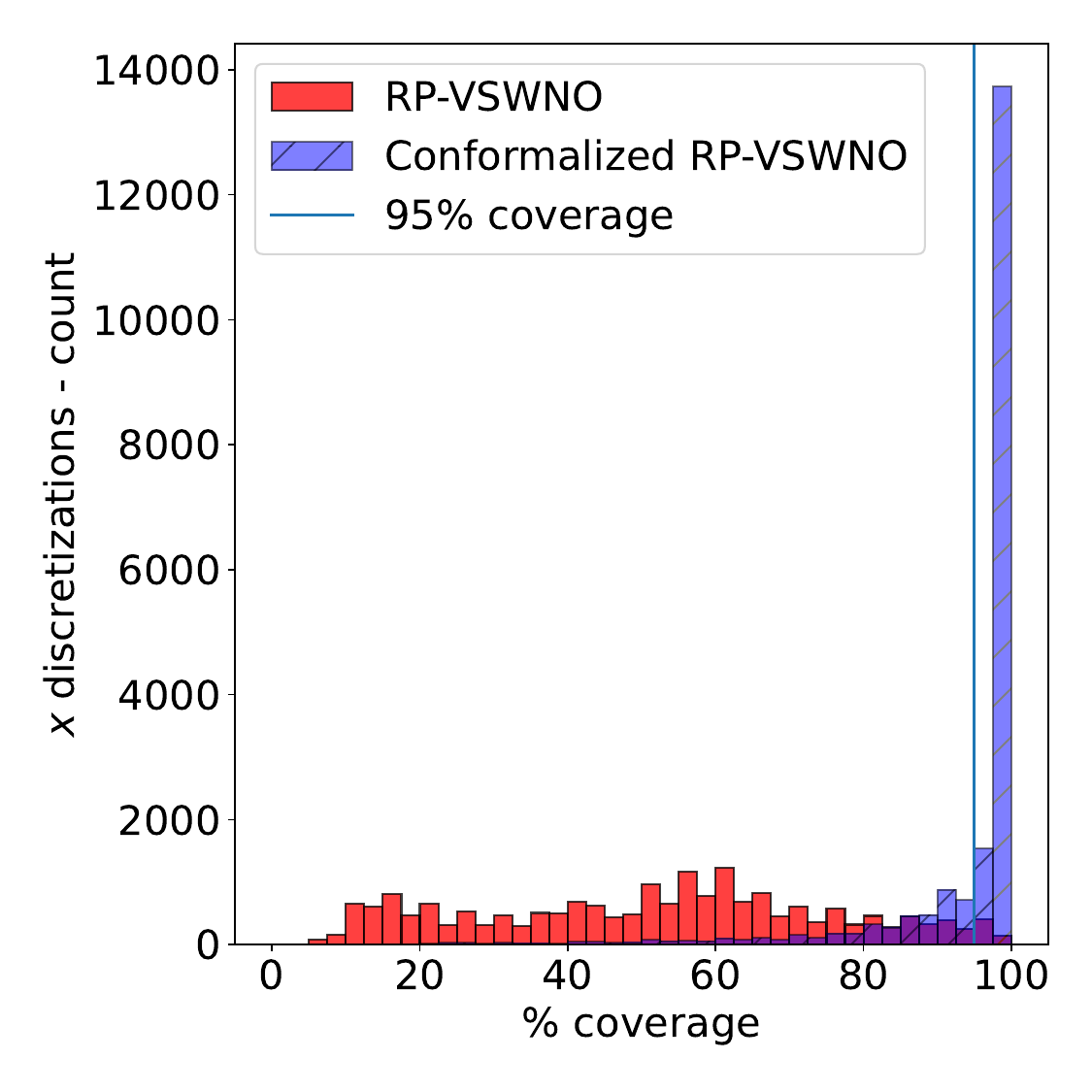}
    \caption{Coverage provided by the confidence intervals.}
    \end{subfigure}\vspace{5mm}
    \begin{subfigure}{0.9\textwidth}
    \centering
    \includegraphics[width=\textwidth]{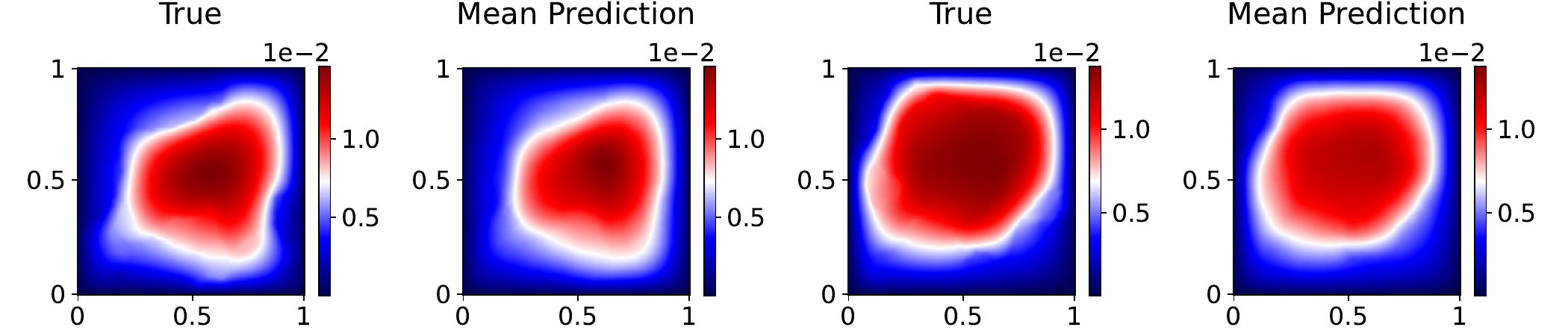}
    \caption{Mean predictions obtained using RP-VSWNO compared against ground truth. NMSE = 0.849\%.}
    \end{subfigure}
    \caption{Super-resolution results for E-II.}
    \label{fig:DarcyRectSuper}
\end{figure}
Figure \ref{fig:DarcyRectSuper} illustrates the super-resolution results using RP-VSWNO as the base model. The training and calibration grid resolution was set to \(85 \times 85\), while the prediction grid resolution was increased to \(141 \times 141\). During calibration, the parameter \(\bm q \in \mathbb{R}^{85 \times 85}\) was computed and mapped to the \(85 \times 85\) grid using a Gaussian process. Subsequently, \(\bm q\) was predicted for the refined \(141 \times 141\) grid.
Figure \ref{fig:DarcyRectSuper}(a) demonstrates that the confidence intervals for super-resolution predictions improve significantly, as opposed to randomized prior operator, when employing CRP-VSWNO. Figure \ref{fig:DarcyRectSuper}(b) compares the predictions from CRP-VSWNO against the ground truth for the higher resolution grid. The results indicate that the model provides a reasonable approximation of the ground truth.

\subsection{E-III, Darcy on Triangular Domain with a Notch}
This example again deals with a two dimensional Darcy equation, defined as,
\begin{equation}
    -\nabla(0.1\,\nabla u(x,y)) = -1,\,\,x,y\in[0,1].
\end{equation}
However, the equation is now defined on a triangular domain $\Omega$ with a rectangular notch. The boundary condition $u(x,y)|_{\partial \Omega}$ is assumed to be random, and sampled from a Gausian random field,
\begin{equation}
\begin{gathered}
    u(\bm x) = GP(0,K(\bm x,\bm x')),\\
    K(\bm x, \bm x') = \exp\left(-\dfrac{1}{2l^2}(\bm x-\bm x')^T(\bm x-\bm x')\right),\,\,l=0.2,\,\,\bm x,\bm x' \in [0,1]^2,
\end{gathered}
\end{equation}
where $\bm x$ and $\bm x'$ represent different grid points on the boundary. In this example, the objective is to learn the operator that maps the boundary conditions to the solution $u(x,y)$,
\begin{equation}
    \mathcal M: \left.u(x,y)\right|_{\delta \Omega} \mapsto u(x,y)
\end{equation}
For further details, interested readers can refer \cite{lu2022comprehensive}.

\begin{figure}[ht!]
    \centering
    \begin{subfigure}{0.9\textwidth}
    \centering
    \includegraphics[width=\textwidth]{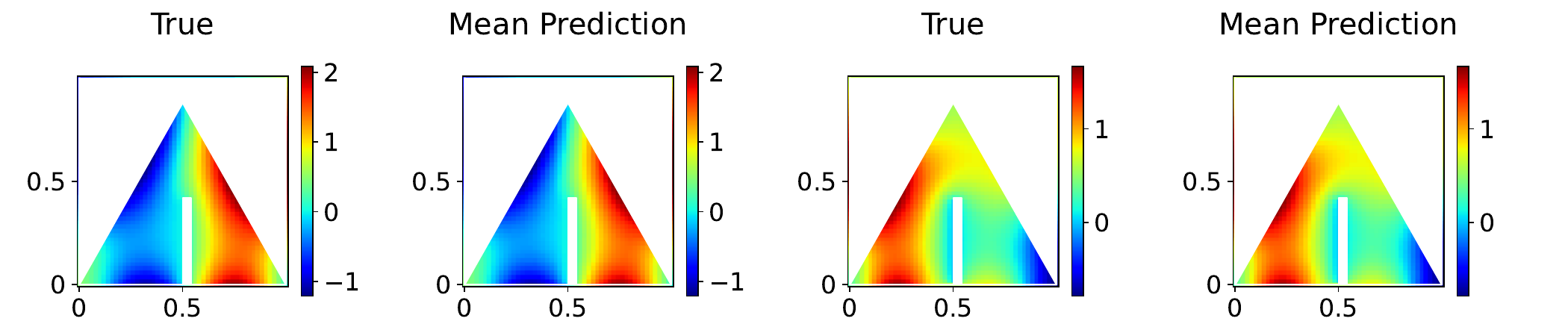}
    \caption{CRP-WNO, NMSE = 0.002\%.}
    \end{subfigure}
    \begin{subfigure}{0.9\textwidth}
    \centering
    \includegraphics[width=\textwidth]{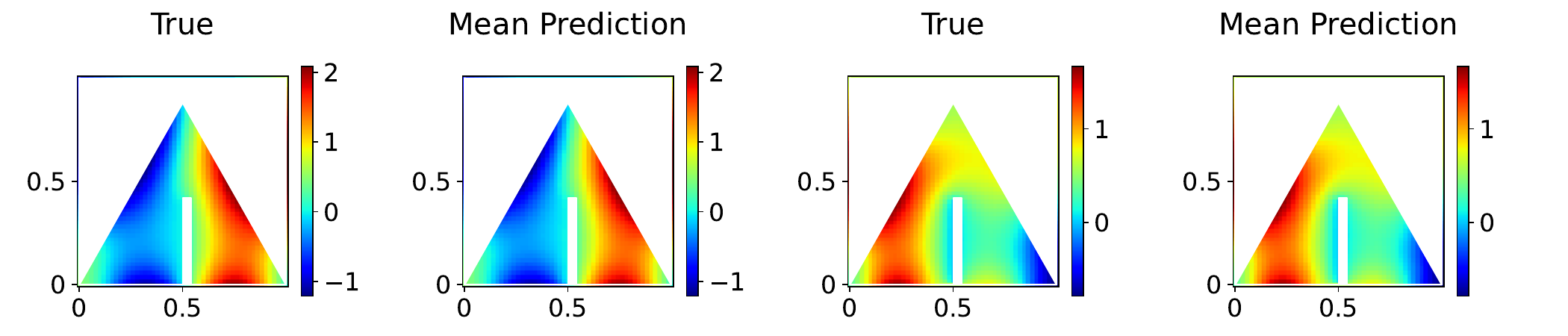}
    \caption{CRP-VSWNO, NMSE = 0.005\%}
    \end{subfigure}
    \caption{Mean predictions for test inputs obtained using CRP-WNO and CRP-VSWNO compared against the ground truth in E-III.}
    \label{fig:DarcyRectPred}
\end{figure}
\begin{table}[ht!]
    \centering
    \caption{Coverage provided by the confidence intervals generated using various deep learning models and their conformalized versions in E-III. Count for the number of discretizations of $(x,y)$ grid where coverage is $< 95\%$ or is $\geq 95\%$ is also included.}
    \vspace{0.5em}
    \label{tab:DarcyRectCoverage}
    \begin{tabular}{lccccc}
    \toprule
    \multirow{2}{*}{Model} & \multicolumn{3}{c}{Coverage} & \multicolumn{2}{c}{Count (Out of 1102)} \\
    \cmidrule(l{2pt}r{2pt}){2-4}\cmidrule(l{2pt}r{2pt}){5-6}
    & Average & Min. & Max. & $<$ 95\% & $\geq$ 95\%\\
    \midrule
    RP-WNO& 94.89 & 0.00 & 100.00 & 141 & 961\\
    \textbf{CRP-WNO} & \textbf{99.70} & \textbf{95.00} & \textbf{100.00} & \textbf{0} & \textbf{1102}\\
    RP-VSWNO & 97.18 & 2.00 & 100.00 & 59 & 1043\\
    \textbf{CRP-VSWNO} & \textbf{99.75} & \textbf{97.00} & \textbf{100.00} & \textbf{0} & \textbf{1102}\\
    Q-WNO & 21.20 & 0.00 & 66.00 & 1102 & 0\\
    CQ-WNO & 94.33 & 82.00 & 100.00 & 515 & 587\\
    \bottomrule
     & 
    \end{tabular}
\end{table}
Results shown in Fig. \ref{fig:DarcyRectPred} and Table \ref{tab:DarcyRectCoverage} reinforce the trends observed in previous examples. CRP-WNO and CRP-VSWNO are able to give expected coverage, and at the same time, the mean predictions obtained closely follow the ground truth.
\subsection{E-IV, Helmholtz Equation}
The fourth example deals with the Helmholtz equation, which is a time-independent form of the wave equation, and is often used to study the effect of traveling waves in areas like seismic ground exploration, radiation studies, etc. The governing equation in this case is defined as follows, 
\begin{equation}
    \nabla^2u(x,z;\omega)+\dfrac{\omega^2}{\nu(x,z)^2}u(x,z;\omega) = f(x_s,z_s;\omega),\,\,x,z\in[0,690],\,\,\delta x,\delta z = 70
\end{equation}
where $u(x,z;w)$ is the magnitude of frequency domain pressure wavefield, corresponding to angular frequency $\omega$ at location $(x,z)$. $(x_s,z_s)$ are the coordinates of the point source, $\nu(x,z)$ is the velocity at location $(x,z)$. While generating the dataset, the Helmholtz equation is not solved directly; rather, the two-dimensional wave equation is solved first in the time domain, and then the solution is converted to the frequency domain to obtain the solution of the Helmholtz equation. The wave equation is defined as,
\begin{equation}
    \nabla^2p(x,z,t)-\dfrac{1}{\nu(x,z)^2}\dfrac{\partial^2}{\partial t^2}p(x,z,t) = s(x_s,z_s,t),
\end{equation}
where $p(x,z,t)$ is the pressure wavefield and $s(x_s,z_s,t)$ is the point source located at $(x_s,z_s)$. While solving, the values for various parameters are considered as follows: $x_s = 680$, $z_s = 10$, and $s(x_s,z_s,t) = 1$. Mur's absorbing boundary conditions are considered, and a spatial resolution of $70\times 70$ is considered. The velocity of the wave over the domain creates a distinction between different samples of the dataset, and the various velocity maps used are taken from the OpenFWI datasets\footnote{https://openfwi-lanl.github.io/docs/data.html, accessed on 12-11-2024, FlatVelA dataset} \cite{deng2022openfwi}.

The Helmholtz equation can be solved across a wide range of frequencies, but lower frequencies typically dominate the solution's behavior. In this example, we focus on frequencies \(\leq 10\) Hz, specifically 1 Hz, 3 Hz, 5 Hz, 7 Hz, and 9 Hz. The model maps the input, comprising angular frequency information and the velocity field \(\nu(x, z)\), to the corresponding output \(u(x, z; \omega)\),
\begin{equation}
    \mathcal M : \nu(x, z) \mapsto u(x, z; \omega),
\end{equation}
where
the output \(u(x, z; \omega)\) consists of both real and imaginary components, necessitating the use of two separate models to capture the complete solution.
We provide frequency information as input to the network, and hence, a single neural operator can capture the response corresponding to all frequencies. Given the complexity of this problem, we consider 20 copies for the randomized prior network and select the best 10.


\begin{table}[ht!]
    \centering
    \caption{Percentage NMSE values observed in CRP-WNO and CRP-VSWNO networks of E-IV, when comparing network's mean predictions with ground truth.}
    \vspace{0.5em}
    \label{tab:HelmholtzNMSE}
    \begin{tabular}{ccccccc}
    \toprule
    \multirow{2}{*}{Network} & \multirow{2}{*}{Component} & \multicolumn{5}{c}{Frequency} \\
    \cmidrule{3-7}
    && 1Hz & 3Hz & 5Hz & 7Hz & 9Hz\\
    \midrule
    \multirow{2}{*}{CRP-WNO} & Real & 0.024 & 0.024 & 0.075 & 0.184 & 0.519 \\
    & Imaginary & 0.002 & 0.002 & 0.004 & 0.006 & 0.011\\ 
    \multirow{2}{*}{CRP-VSWNO} & Real & 0.047 & 0.069 & 0.232 & 0.615 & 1.297 \\
    & Imaginary & 0.004 & 0.008 & 0.011 & 0.018 & 0.029\\
    \bottomrule
    \end{tabular}
\end{table}
Table \ref{tab:HelmholtzNMSE} shows the NMSE values obtained when comparing the mean predictions obtained using CRP-WNO and CRP-VSWNO models against the corresponding ground truth. As can be seen, the error observed is marginal, and the mean predictions are a good approximation of the ground truth.
\begin{figure}[ht!]
\centering
\begin{subfigure}{1\textwidth}
    \centering
    \includegraphics[width=0.95\linewidth]{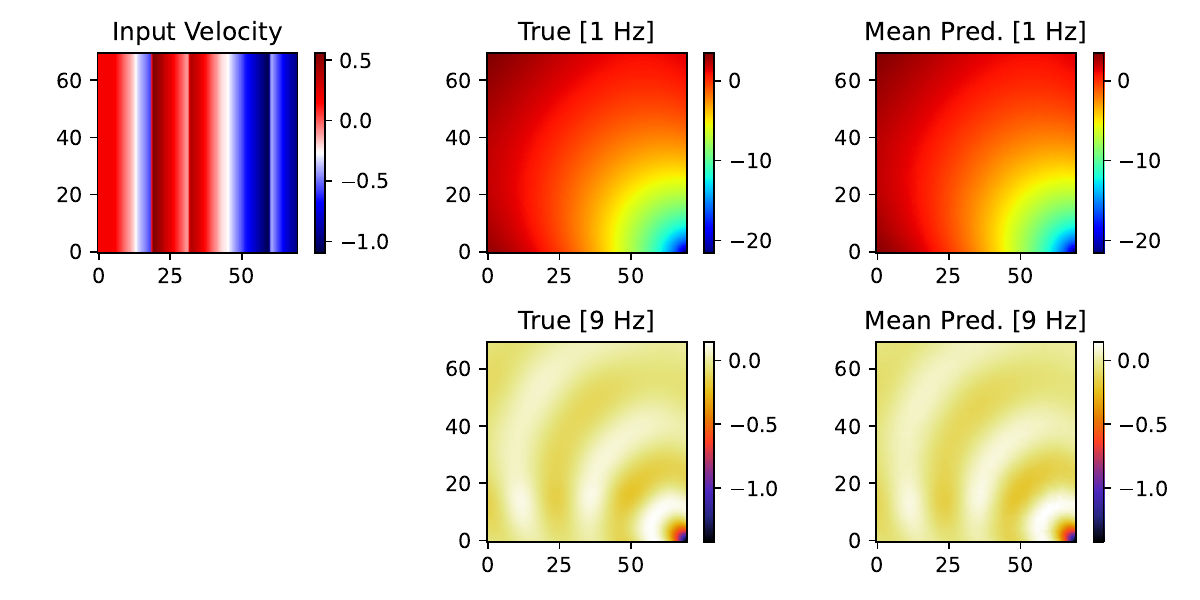}
    \caption{Real Component}
    \label{fig:real_1hz9hz}
\end{subfigure}
\begin{subfigure}{1\textwidth}
    \centering
    \includegraphics[width=0.95\linewidth]{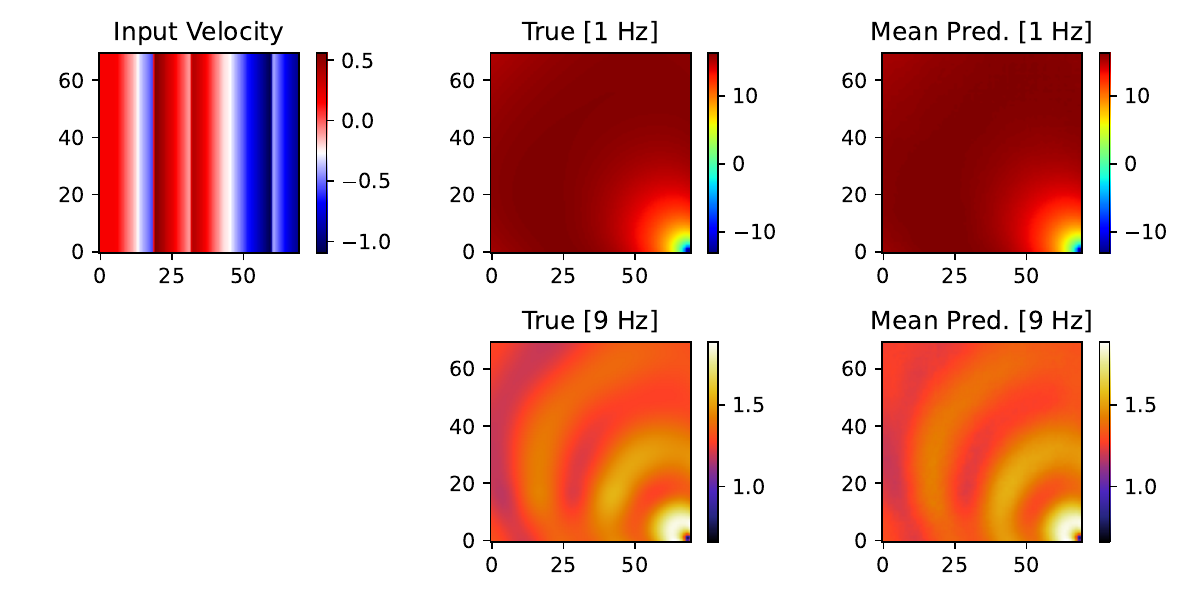}
    \caption{Imaginary Component}
    \label{fig:imag_1hz9hz}
\end{subfigure}
\caption{Mean predictions corresponding to one test input obtained using CRP-VSWNO compared against the ground truth in E-IV.}
\label{fig:real_imag_1hz9hz}
\end{figure}
Fig. \ref{fig:real_imag_1hz9hz} compares the mean prediction of CRP-VSWNO the ground truth corresponding to one test input. Both real and imaginary components corresponding to 1Hz and 9Hz frequencies are plotted. As can be seen, the mean prediction closely follows the ground truth.
\begin{table}[ht!]
\centering
\caption{Count for the number of discretizations of (x, y) grid where coverage is $<$95\% or is $\geq$95\% obtained using various deep learning models and their conformalized versions in E-IV.}
\label{tab:HelmholtzCount}
\begin{subtable}{0.9\textwidth}
\caption{Results corresponding to $\omega = 1$Hz.}
\resizebox{\textwidth}{!}{
\begin{tabular}{lccc>{\bfseries}cc>{\bfseries}ccc}
    \toprule
    \multicolumn{3}{l}{Model} & RP-WNO & CRP-WNO & RP-VSWNO & CRP-VSWNO & Q-WNO & CQ-WNO\\
    \midrule
    \multirow{4}{*}{\parbox{1.250cm}{Count (out of 4900)}} & \multirow{2}{*}{$<95\%$}& Real & 2589&0&2496&0&4900&3594\\
    &&Imaginary& 2644&0&2858&0&4697&1705\\
    &\multirow{2}{*}{$\geq95\%$}& Real&2311&4900&2404&4900&0&1306\\
    &&Imaginary& 2256&4900&2042&4900&203&3195\\
    \bottomrule
     & 
    \end{tabular}
    }
\end{subtable}
\begin{subtable}{0.9\textwidth}
\caption{Results corresponding to $\omega = 3$Hz.}
\resizebox{\textwidth}{!}{
\begin{tabular}{lccc>{\bfseries}cc>{\bfseries}ccc}
    \toprule
    \multicolumn{3}{l}{Model} & RP-WNO & CRP-WNO & RP-VSWNO & CRP-VSWNO & Q-WNO & CQ-WNO\\
    \midrule
    \multirow{4}{*}{\parbox{1.250cm}{Count (out of 4900)}} & \multirow{2}{*}{$<95\%$}& Real & 1330&0&958&0&4900&3132\\
    &&Imaginary& 1950&1&2528&0&4897&3016\\
    &\multirow{2}{*}{$\geq95\%$}& Real& 3570&4900&3942&4900&0&1768\\
    &&Imaginary& 2950&4899&2372&4900&3&1884\\
    \bottomrule
     & 
    \end{tabular}
    }
\end{subtable}
\begin{subtable}{0.9\textwidth}
\caption{Results corresponding to $\omega = 5$Hz.}
\resizebox{\textwidth}{!}{
\begin{tabular}{lccc>{\bfseries}cc>{\bfseries}ccc}
    \toprule
    \multicolumn{3}{l}{Model} & RP-WNO & CRP-WNO & RP-VSWNO & CRP-VSWNO & Q-WNO & CQ-WNO\\
    \midrule
    \multirow{4}{*}{\parbox{1.250cm}{Count (out of 4900)}} & \multirow{2}{*}{$<95\%$}& Real & 2700&0&699&0&4900&1746\\
    &&Imaginary& 2943&0&2174&0&4897&1906\\
    &\multirow{2}{*}{$\geq95\%$}& Real& 2200&4900&4201&4900&0&3154\\
    &&Imaginary& 1957&4900&2726&4900&3&2994\\
    \bottomrule
     & 
    \end{tabular}
    }
\end{subtable}
\begin{subtable}{0.9\textwidth}
\caption{Results corresponding to $\omega = 7$Hz.}
\resizebox{\textwidth}{!}{
\begin{tabular}{lccc>{\bfseries}cc>{\bfseries}ccc}
    \toprule
    \multicolumn{3}{l}{Model} & RP-WNO & CRP-WNO & RP-VSWNO & CRP-VSWNO & Q-WNO & CQ-WNO\\
    \midrule
    \multirow{4}{*}{\parbox{1.250cm}{Count (out of 4900)}} & \multirow{2}{*}{$<95\%$}& Real & 3544&0&1097&0&4900&1877\\
    &&Imaginary& 3243&0&2627&0&4900&1731\\
    &\multirow{2}{*}{$\geq95\%$}& Real& 1356&4900&3803&4900&0&3023\\
    &&Imaginary& 1657&4900&2273&4900&0&3169\\
    \bottomrule
     & 
    \end{tabular}
    }
\end{subtable}
\begin{subtable}{0.9\textwidth}
\caption{Results corresponding to $\omega = 9$Hz.}
\resizebox{\textwidth}{!}{
\begin{tabular}{lccc>{\bfseries}cc>{\bfseries}ccc}
    \toprule
    \multicolumn{3}{l}{Model} & RP-WNO & CRP-WNO & RP-VSWNO & CRP-VSWNO & Q-WNO & CQ-WNO\\
    \midrule
    \multirow{4}{*}{\parbox{1.250cm}{Count (out of 4900)}} & \multirow{2}{*}{$<95\%$}& Real & 4382&0&1196&0&4900&1805\\
    &&Imaginary& 3413&0&2755&0&4898&1874\\
    &\multirow{2}{*}{$\geq95\%$}& Real& 518&4900&3704&4900&0&3095\\
    &&Imaginary& 1487&4900&2145&4900&2&3026\\
    \bottomrule
     & 
    \end{tabular}
    }
\end{subtable}
\end{table}
Table \ref{tab:HelmholtzCount} shows the number of discretizations where coverage is $<95\%$ or $\geq 95\%$ when using RP-WNO, RP-VSWNO, Q-WNO or their conformalized variants. It should be noted that although during the training of various deep learning models, the dataset contained information corresponding to all the frequencies, during calibration, outputs corresponding to different frequencies were calibrated individually. This is done to show the flexibility of the adopted approach. 
The results show that barring a single grid location, the expected coverage of $\geq 95\%$ is achieved at almost all locations when using CRP-WNO or CRP-VSWNO, whereas other deep learning models fail to achieve the same.

\begin{table}[ht!]
    \centering
    \caption{Comparison between the performance of RCQNO and CRP-WNO. Since, the average coverage obtained is $\geq 95\%$ in both approaches, we compare using the average width of the calibrated confidence interval $d$ and the percentage NMSE values obtained when comparing predictions against the ground truth. $n$ here represents the quantum of calibration data used.}
    \vspace{0.5em}
    \label{tab:RCQNO comparison}
    \begin{tabular}{llcccccc}
    \toprule
    Component & Frequency & \multicolumn{3}{c}{RCQNO} & \multicolumn{3}{c}{CRP-WNO}
    \\
    &&$n$&$d\,(\times 10^{-2})$&\% NMSE&$n$&$d(\times 10^{-2})$&\% NMSE
    \\
    \midrule
    \multirow{5}{*}{Real}&1Hz&\multirow{5}{*}{500}&48.88&0.023&\multirow{5}{*}{150}&\textbf{33.75}&\textbf{0.024}
    \\
    &3Hz&&22.64&0.204&&\textbf{6.36}&\textbf{0.024}
    \\
    &5Hz&&22.59&0.910&&\textbf{4.97}&\textbf{0.075}
    \\
    &7Hz&&22.28&2.482&&\textbf{4.67}&\textbf{0.184}
    \\
    &9Hz&&23.37&4.178&&\textbf{5.15}&\textbf{0.519}
    \\[0.15em]
    \multirow{5}{*}{Imaginary}&1Hz&\multirow{5}{*}{500}&122.73&0.013&\multirow{5}{*}{150}&\textbf{51.89}&\textbf{0.002}
    \\
    &3Hz&&53.94&0.023&&\textbf{15.21}&\textbf{0.002}
    \\
    &5Hz&&44.51&0.027&&\textbf{13.01}&\textbf{0.004}
    \\
    &7Hz&&38.01&0.041&&\textbf{12.27}&\textbf{0.006}
    \\
    &9Hz&&34.16&0.060&&\textbf{11.96}&\textbf{0.011}
    \\
    \bottomrule
    \end{tabular}
\end{table}
As an additional case study, Table \ref{tab:RCQNO comparison} illustrates  the performance of CRP-WNO as compared to RCQNO. The results obtained for RCQNO are based on the code implementation given in the GitHub\footnote{https://github.com/ziqi-ma/neuraloperator/tree/ziqi/uq/uq, accessed on: 04 November 2024},\footnote{https://github.com/ziqi-ma/neuraloperator/blob/ziqi/uq/uq/FNO\_darcy\_residual.py, accessed on: 04 November 2024} repository. As can be seen from the table, CRP-WNO has produced tighter bounds, and the NMSE observed is also lower for the CRP-WNO network. The RCQNO framework failed to provide the required coverage when predicting imaginary component corresponding to 1Hz and 3Hz frequencies at one and 45 grid locations respectively. On the contrary, the CRP-WNO failed to provide the coverage at a single grid location only, when predicting imaginary component corresponding to 3Hz frequency.
\section{Conclusion}\label{section: conclusion}
The estimates for the uncertainty associated with the neural operator predictions give users auxiliary confidence when undertaking decision-making tasks. These estimates become especially essential for the spiking variants of these operators as they exhibit sparse communication to conserve energy, which can potentially lead to reduced accuracy in predictions. To address this, we propose in this paper, a distribution free uncertainty quantification framework in neuroscience inspired spiking operators.
The proposed framework uses the uncertainty bounds from randomized prior (RP) operators as its initial uncertainty estimates and calibrates them using the split conformal prediction (SCP) algorithm to provide the required coverage. The selection of the RP operator for obtaining the initial estimate of uncertainty is based on the hypothesis that a good initial estimate will improve the SCP algorithm's calibration performance. It is also based on the fact that the RP operators are easily scalable to complex architectures, are computationally efficient, and can account for prior information while being set in a deterministic framework. 

Super-resolution is a defining property for operator learning algorithms, and gauging uncertainty associated with super-resolution predictions is important because, in this case, the trained model is predicting at solution grid locations that are unseen during training. The conventional SCP framework, unfortunately, is not designed for zero-shot super-resolution. 
In response to this challenge, we also enhance the algorithm to allow uncertainty quantification in zero-shot super-resolution setup. The developed algorithm is integrated with the recently developed variable spiking wavelet neural operator (VSWNO) and the vanilla wavelet neural operator (WNO) so as to enable uncertainty quantification in these operators.  The resulting algorithms are referred to as CRP-VSWNO and CRP-WNO.
%
%
%

To test the validity of the proposed framework, we show four examples covering both one-dimensional and two-dimensional partial differential equations. VSWNO and vanilla WNO are used as our base operator learning algorithms. The key findings of the paper are as follows,
\begin{itemize}
    \item With the exception of example example four, where the coverage criterion is not satisfied at one of the location of the solution grid, the CRP-WNO and the CRP-VSWNO are able to give the required coverage at all locations of the solution grid when tested across other examples. On the contrary, the conformalized quantile variant fails to yield desired coverage at several locations. This establishes the importance of initial uncertainty estimates as hypothesized in this work.
    \item As illustrated in the first two example, the proposed approach yield excellent results in the zero-shot super-resolution scenario as well. This illustrates that the efficacy of the proposed zero-shot super-resolution enabled split conformal algorithm. 
    \item In the fourth example, we also compare the results obtained using the proposed framework with those obtained using the risk controlling quantile neural operator. While both the frameworks were yield desired coverage on average (over all locations of the solution grid), the proposed approach yield tighter bounds, and its mean prediction gave a better reflection of the ground truth.
    Also, unlike the proposed approach, which produces a separate conformal parameter for all locations of the solution grid, the RCQNO framework produces a single conformal parameter for the whole domain, leading to a possible overestimation of uncertainty at some locations on the output domain.
\end{itemize}
Overall, the proposed framework benefits from the good initial predictions of the RP operators, and subsequently, the SCP can calibrate them with a meager dataset. The GP also aids in improving the uncertainty estimates compared to the initial RP-operator estimates in a super-resolution setting. To further extend the literature on the discussed framework, studies concerning the optimum size of the calibration dataset may be carried out. Furthermore, a comprehensive empirical study showing whether the conditional coverage is obtained when using the discussed framework may also be carried out. 
\section*{Acknowledgment}
SG acknowledges the financial support received from the Ministry of Education, India, in the form of the Prime Minister's Research Fellows (PMRF) scholarship. SC acknowledges the financial support received from Anusandhan National Research Foundation (ANRF) via grant no. CRG/2023/007667 and from the Ministry of Port and Shipping via letter no. ST-14011/74/MT (356529). 


\appendix
\section{Wavelet Neural Operator and the Architecture Details}\label{appendix:arch}
In this section, we discuss the architecture of WNO based deep learning models used in various examples. But before delving into its architecture, let us first discuss the theory behind theory behind neural operators and WNO. Neural operators are a class of deep learning architecture that aims to learn some operator $\mathcal O_P:\mathcal I \rightarrow \mathcal O$, that maps the input function space $\mathcal I$ to the output function space $\mathcal O$. To achieve this, a parametric map $\mathcal G_\phi$ is formed that is described as follows,
\begin{equation}
    \mathcal O_P\simeq\mathcal G_\phi=\mathcal P \circ \sigma(\mathcal W_L+\mathcal K_L+b_L)\circ\cdots\circ \sigma(\mathcal W_1+\mathcal K_1+b_1)\circ\mathcal U,
\end{equation}
where $\mathcal P$ is the projecting operator, $\mathcal U$ is the uplifting operator, $\mathcal W_i$ is a local operator, $\mathcal K_i$ is a kernel integral operator and $b_i$ is a bias function. The operators $\mathcal U$ and $\mathcal P$ are usually represented in neural operators by a simple neural network, and $\mathcal W_i$ is usually represented by a pointwise neural network, generally a convolution layer. $\sigma(\cdot)$ represents a continuous activation and the operation $\sigma(\mathcal W_i+ \mathcal K_i + b_i)$ is repeated for $L$ layers. The definition of the integral operator $\mathcal K_i$ is unique to the specific operator learning algorithm in consideration. In WNO, for an arbitrary function $\bm v_i$ with domain $\mathcal D$, which is input to the $i$\textsuperscript{th} intermediary layer, the operator $\mathcal K_i$ is defined as,
\begin{equation}
    \mathcal K_iv_i(x) = W^{-1}(R_iW(v_i))(x), \hspace{2.5em}x\in\mathcal D,
    \label{eq:itrlayer}
\end{equation}
where $W(\cdot)$ and $W^{-1}(\cdot)$ represent the wavelet and inverse wavelet transform respectively and $R_i$ is a parameterized kernel. The wavelet and inverse wavelet transform operations on a function $f:\mathcal D\rightarrow\mathbb R^{d_f}$ is defined as,
\begin{equation}
\begin{gathered}
    W(f)(x) = f_w(\gamma,\beta) = \int_{\mathcal D}f(x)\dfrac{1}{|\gamma|^{1/2}}\psi\left(\dfrac{x-\beta}{\gamma}\right)dx,\\[0.65em]
    W^{-1}(f_w)(\gamma,\beta)=f(x)=\dfrac{1}{C_\psi}\int_0^\infty\int_{\mathcal D}f_w(\gamma,\beta)\dfrac{1}{|\gamma|^{1/2}}\widetilde\psi\left(\dfrac{x-\beta}{\gamma}\right)d\beta \dfrac{d\gamma}{\gamma^2},\\[0.65em]
    C_\psi = 2\pi\int_{\mathcal D}\dfrac{|\psi_w|^2}{w}
\end{gathered}
\end{equation}
where $\gamma\in\mathbb R^+$ and $\beta\in\mathbb R$ are the scaling and translation parameters respectively. $\psi(\cdot)$ is an orthonormal mother wavelet and $\widetilde\psi(\cdot)$ is its dual function. $\psi_w$ represents the Fourier transform of the mother wavelet $\psi$. For more details regarding WNO architecture, readers are advised to follow \cite{tripura2023wavelet}.

The model selection and hyperparameter selection for WNO models in various examples were done based on random search optimization and experience from working on WNO architecture. ADAM optimizer was used to optimize various deep learning models, and the learning rate was taken to be equal to 0.001. $L^2$ error was used to train the RP-WNO and RP-VSWNO networks. In RP-VSWNO networks, the spiking activity of VSNs was also penalized using SLF. The projecting operator $\mathcal P$ and uplifting operator $\mathcal U$ are modeled using two-layered and single-layered neural networks. In the neural network representing the projecting operator, the final layer has a single node in all the examples. The remaining details regarding network architecture are given in Table \ref{tab:networkdetails}.
\begin{table}[ht!]
    \centering
    \caption{Network architecture details in various examples. CWT refers to Continuous Wavelet Transform and DWT refers to Discrete Wavelet Transform. $L$ refers to the number of iterative layers (refer \eqref{eq:itrlayer}) used, and $m$ refers to the number of wavelet decompositions carried out in each iterative layer.}
    \vspace{0.5em}
    \label{tab:networkdetails}
    \begin{tabular}{lccccc}
        \toprule
        \multirow{2}{*}{Example} & \multicolumn{2}{c}{Nodes in first layer} & \multirow{2}{*}{$L$} & \multirow{2}{*}{Wavelet} & \multirow{2}{*}{$m$}\\\cmidrule{2-3}
        &$\mathcal P$& $\mathcal U$&\\
        \midrule
        E-I, Burgers Equation & 128 & 64 & 4 & DWT - db6 & 8\\
        E-II, Darcy Equation on Rectangular Domain & 128 & 64 & 4 & DWT - db4& 4\\        
        E-III, Darcy Equation on Triangular Domain & 128 & 64 & 4 & DWT - db6 & 3\\        
        E-IV, Helmholtz Equation - Real Component & 64 & 64 & 4 & CWT & 4\\
        \bottomrule
    \end{tabular}
\end{table}
GeLU activation is used between two iterative layers and between two layers of the network representing the projecting operator $\mathcal P$. Inputs and outputs are normalized in E-II to E-IV. In the prior network of the RP networks, all other details are kept the same, except the number of iterative layers is reduced to two. In VS-WNO networks, the continuous activations are replaced by VSNs in the trainable network of RP setup. Any other miscellaneous details can be referred from the sample codes provided in the GitHub repository (to be released after acceptance). 
\section{Q-WNO and CQ-WNO}\label{appendix:qwno}
The Quantile WNO or Q-WNO, as referenced in the text, is a variant of vanilla WNO, which is trained using the concepts of quantile regression \cite{hao2007quantile,koenker2001quantile,moya2024conformalized}. The idea here is to train two separate models, one corresponding to $\alpha/2$ quantile and the other corresponding to $1-\alpha/2$ quantile. The respective predictions obtained combined should, in theory, give coverage such that the true outputs lie between the two predictions with probability $\geq (1-\alpha)$. Now to train the vanilla WNO network corresponding to $\eta$ quantile, the loss function $L$ is defined as follows,
\begin{equation}
    L = \left\{\begin{array}{cl}
        \eta ||\bm y-\widetilde{\bm y}||, & \text{if }||\bm y||\geq ||\widetilde{\bm y}|| \\
        (1-\eta) ||\bm y-\widetilde{\bm y}||, & \text{if } ||\bm y||< ||\widetilde{\bm y}||
    \end{array}\right.,
\end{equation}
where $\bm y$ is the true solution and $\widetilde{\bm y}$ is the corresponding prediction. Now, let $\widetilde{\bm y_1}$ be the prediction corresponding to the network trained for $\eta=\alpha/2$ and $\widetilde{\bm y_2}$ be the prediction corresponding to the network trained for $\eta=1-\alpha/2$. The initial confidence interval can thus be defined as,
\begin{equation}
    \mathcal C_{ini} = [\widetilde{\bm y_1},\,\,\widetilde{\bm y_2}].
\end{equation}
The calibrated confidence intervals for CQ-WNO can then be obtained as, 
\begin{equation}
    \mathcal C_{p} = [\widetilde{\bm y_1}-\bm q,\,\,\widetilde{\bm y_2}+\bm q].
\end{equation}
$\bm q$ is computed using a procedure similar to the one explained in Algorithm \ref{algo} with the exception that the base model is now Q-WNO and the score function corresponding to $j$\textsuperscript{th} element $y^{\,(j)}$ of the output vector $\bm y$ is computed as,
\begin{equation}
    e^{(j)} = max\{\widetilde y_1^{\,(j)}-y^{(j)}, y^{(j)}-\widetilde y_2^{\,(j)}\},
\end{equation}
where $y_1^{\,(j)}$ and $y_1^{\,(j)}$ are the $j$\textsuperscript{th} elements of $y_1^{\,(j)}$ and $y_2^{\,(j)}$ respectively.
\section{Spiking Activity}\label{appendix:spkact}
In the various examples, the CRP-VSWNO networks are formed by using four VSNs in place of four continuous activations of CRP-WNO networks. Now to gauge the energy efficiency of VSNs, we report their spiking activity in Table \ref{tab:spkactvsn}. The idea here is that in event-driven hardware, reduced neuron activity will lead to reduced computations and, hence, more energy saving. For example, \cite{garg2023neuroscience,garg2024neuroscience} discuss that for VSNs to be energy efficient (compared to when using continuous activations) in synaptic operations of convolution layers, their spiking activity has to be less than 100\%. It should be noted that in order to increase sparsity in communication during training, the spiking activity of the VSNs can also be penalized by incorporating the same in the loss function. The modified loss function $L_s$ is termed as the Spiking Loss Function (SLF) and is defined as,
\begin{equation}
    L_s = \alpha_w L_v + \beta_w S_a,
\end{equation}
where $L_v$ is the vanilla loss function like mean squared error and $S_a$ is the total spiking activity ratio of all the VSNs, whose spiking activity needs to be constrained. The parameters $\alpha_w$ and $\beta_w$ are the weights assigned to each loss function and the same act as hyper-parameters for the operator network. This can lead to marginal decrease in accuracy but the same can be managed by adopting appropriate weights in the loss function.
\begin{table}[H]
    \centering
    \caption{Average spiking activity observed for various VSNs placed in RP-VSWNO networks. The reported spiking activity is calculated as the 100 $\times$ number of spikes observed/ total possible spikes. The spiking activity is averaged over all samples of the test dataset and all models of the RP ensemble.}
    \vspace{0.5em}
    \label{tab:spkactvsn}
    \begin{tabular}{lcccc}
        \toprule
        \multirow{2}{*}{Example} & \multicolumn{4}{c}{VSN}\\\cmidrule{2-5}
        & 1 & 2 & 3 & 4\\
        \midrule
        E-I& 30.27 & 22.90	& 16.16	& 3.05\\
        E-II& 44.87 & 33.04	& 10.19	& 5.88\\
        E-III& 34.87 & 38.03 & 23.84 & 4.90\\
        E-IV (Real Component Network)& 22.17 & 25.57 & 25.71 & 64.76\\
        E-IV (Imaginary Component Network)& 22.92 & 25.09 & 18.74 & 44.82\\
        \bottomrule
    \end{tabular}
\end{table}
\end{document}